\documentclass{article}

\usepackage{color,xcolor}
\usepackage{epsfig}
\usepackage{graphicx}

\usepackage{adjustbox}
\usepackage{array}
\usepackage{booktabs}
\usepackage{colortbl}
\usepackage{float,wrapfig}
\usepackage{hhline}
\usepackage{multirow}
\usepackage{tabularx}
\usepackage{makecell}
\usepackage{float}

\usepackage{amsmath,amsfonts,amsthm,amssymb}
\usepackage{bm}
\usepackage{nicefrac}
\usepackage{microtype}
\usepackage{url}

\usepackage{changepage}
\usepackage{extramarks}
\usepackage{fancyhdr}
\usepackage{lastpage}
\usepackage{setspace}
\usepackage{soul}

\usepackage{algorithm}
\usepackage{algorithmicx}
\usepackage{algpseudocodex}
\usepackage{enumerate}
\usepackage{enumitem}
\usepackage{todonotes}

\usepackage{etoolbox,siunitx}
\usepackage{afterpage}
\usepackage{subcaption}
\usepackage[title]{appendix}
\usepackage{overpic}

\usepackage[preprint]{neurips_2025}
\usepackage{bbding}

\usepackage[utf8]{inputenc} 
\usepackage[T1]{fontenc}    
\usepackage{hyperref}       
\usepackage{url}            
\usepackage{booktabs}       
\usepackage{amsfonts}       
\usepackage{nicefrac}       
\usepackage{microtype}      
\usepackage{xcolor}         
\usepackage{xpatch}
\usepackage{subcaption}

\usepackage{bm}

\makeatletter
\xapptocmd{\NAT@bibsetnum}{\setlength{\leftmargin}{0pt}\setlength{\itemindent}{\labelwidth}\addtolength{\itemindent}{\labelsep}}{}{}
\makeatother

\newcommand{\mliu}[2]{\textcolor[rgb]{1,0,1}{{\if 0 #1\fi}{#2}}}

\usepackage[capitalize]{cleveref}
\crefname{section}{Sec.}{Secs.}
\Crefname{section}{Section}{Sections}
\Crefname{table}{Table}{Tables}
\crefname{table}{Tab.}{Tabs.}
\Crefname{equation}{Equation}{Equations}
\crefname{equation}{Eqn.}{Eqns.}

\RequirePackage{xspace}

\makeatletter
\DeclareRobustCommand\onedot{\futurelet\@let@token\@onedot}
\def\@onedot{\ifx\@let@token.\else.\null\fi\xspace}

\def\ie{\emph{i.e}\onedot}

\makeatother

\title{GLaVE-Cap: Global-Local Aligned Video Captioning with Vision Expert Integration}
\author{%
{Wan Xu$^{1}$\thanks{*Equal contribution. $\dag$Corresponding author.} ~ Feng Zhu$^{1*}$ ~ Yihan Zeng{$^{2}$} ~ Yuanfan Guo{$^{2}$}} \\ \textbf{Ming Liu{$^{1}$} ~ Hang Xu{$^{2}$} ~ Wangmeng Zuo{$^{1\dag}$}} \\
\normalsize
\vspace{2pt}
$^{1}$\	Harbin Institute of Technology  
$^{2}$\ Huawei Noah’s Ark Lab
}

\begin{document}

\maketitle

\begin{abstract}
Video detailed captioning aims to generate comprehensive video descriptions to facilitate video understanding. Recently, most efforts in the video detailed captioning community have been made towards a local-to-global paradigm, which first generates local captions from video clips and then summarizes them into a global caption. However, we find this paradigm leads to less detailed and contextual-inconsistent captions, which can be attributed to (1) no mechanism to ensure fine-grained captions, and (2) weak interaction between local and global captions.
To remedy the above two issues, we propose \textbf{GLaVE-Cap}, a \textbf{G}lobal-\textbf{L}ocal \textbf{a}ligned framework with \textbf{V}ision \textbf{E}xpert integration for \textbf{Cap}tioning, which consists of two core modules: TrackFusion enables comprehensive local caption generation, by leveraging vision experts to acquire cross-frame visual prompts, coupled with a dual-stream structure; while CaptionBridge establishes a local-global interaction, by using global context to guide local captioning, and adaptively summarizing local captions into a coherent global caption.
Besides, we construct \textbf{GLaVE-Bench}, a comprehensive video captioning benchmark featuring 5$\times$ more queries per video than existing benchmarks, covering diverse visual dimensions to facilitate reliable evaluation.
We further provide a training dataset \textbf{GLaVE-1.2M} containing 16K high-quality fine-grained video captions and 1.2M related question-answer pairs. Extensive experiments on four benchmarks show that our GLaVE-Cap achieves state-of-the-art performance. Besides, the ablation studies and student model analyses further validate the effectiveness of the proposed modules and the contribution of GLaVE-1.2M to the video understanding community. The source code, model weights, benchmark, and dataset will be open-sourced.
 
\end{abstract}
\section{Introduction}
The evolution of vision-language models (VLMs) has marked a significant milestone in multi-modal learning, enabling the joint understanding of visual and textual data~\cite{CLIP, li2023blip, gpt4o}. While early VLMs primarily focused on static image-text tasks, recent efforts have extended this paradigm to the video domain, giving rise to video-language models (Video-LLMs)~\cite{llava_video, sharegpt4video, vript, georgiev2024gemini, bai2025qwen2}. These models are designed to process and reason over temporally extended visual content in conjunction with language, introducing new challenges and opportunities in video understanding. Drawing on insights from the image domain, where fine-grained captions have proven crucial for enhancing visual understanding~\cite{Yan2024FIHAAH,Onoe2024DOCCIDO}, recent research has increasingly focused on constructing fine-grained video caption datasets~\cite{auroracap,sharegpt4video,vript,llava_video}. These fine-grained video captions, which comprehensively describe attributes and motions of all objects and maintain consistency across descriptions, serve as a cornerstone for fine-grained video understanding~\cite{auroracap,sharegpt4video,llava_video}.
\begin{figure}[t]
  \centering
  \includegraphics[width=1\textwidth]{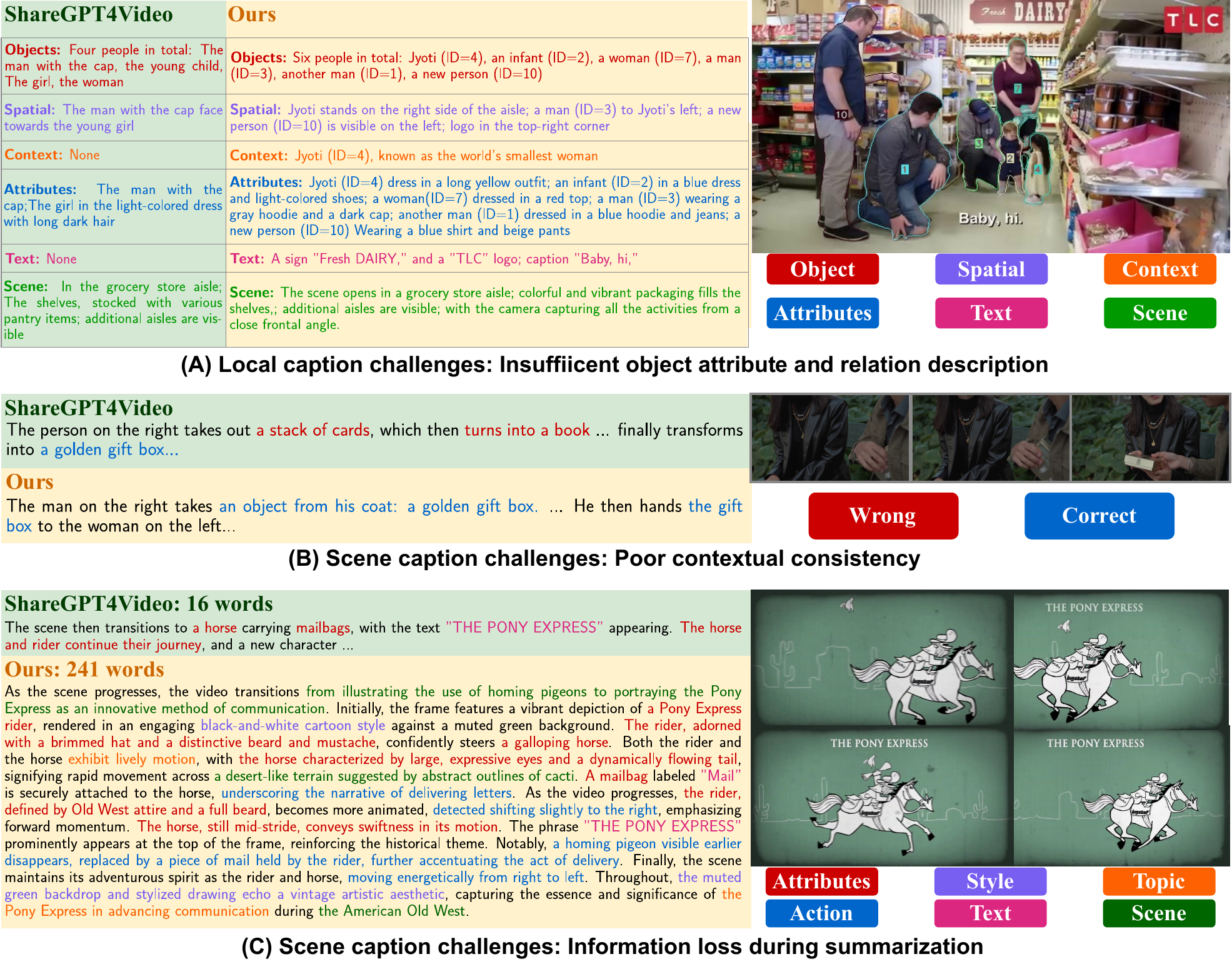}
  \vspace{-1.5em}
  \caption{Illustration of the limitations of current methods. Refer to \cref{fig:full_caption} for full caption of (A).}
  \label{fig:teaser}
  \vspace{-1.0em}
\end{figure}
However, generating detailed video captions remains a significant challenge. Directly applying VLMs to video inputs often leads to the omission of critical visual details~\cite{sharegpt4video}. A common strategy adopted in recent works is the \textbf{local-to-global captioning paradigm}, which first generates local captions for short clips or keyframes and subsequently summarizes them into a holistic video-level description~\cite{lin2023mm, vript,llava_video,sharegpt4video}. This paradigm decomposes the captioning task, allowing VLMs to process only a small set of visual inputs at one time, helping preserve fine-grained details and improving scalability to longer videos.
However, when the local-to-global scheme is applied to complex videos involving multiple objects and scenes, these methods commonly face three limitations, as illustrated in \cref{fig:teaser}: (1) \textbf{Insufficient object attribute and relation description}: Local captions often fail to capture challenging attributes such as object spatial relations and quantity, and they tend to be biased toward dynamic changes while overlooking detailed information about object attributes and the scene context. (2) \textbf{Poor contextual consistency}: Local captions generated based only on partial video content often struggle to accurately interpret visual elements, and may lead to inconsistencies among multiple local captions, ultimately resulting in a contextually inconsistent video caption.
(3) \textbf{Information loss during summarization}: Summarizing all local captions into a single video caption in one step is challenging to LLMs, which inevitably omit a substantial amount of detailed information. Some methods attempt to address this issue by multi-step summarization~\cite{vript,llava_video}, such as summarizing scene-by-scene. However, they often lack an effective strategy to integrate these intermediate results into a coherent and comprehensive video-level caption. Together, the aforementioned issues hinder the generation of accurate and fine-grained video captions.

We attribute the limitations above to two underlying causes. 
First, local caption generation relies solely on carefully crafted text prompts, which offer limited control over output quality when facing challenging tasks, such as counting. In addition, the sequential nature of keyframe inputs predisposes VLMs to focus on inter-frame dynamics, while static details are frequently overlooked.
Second, the limited interaction between local and global captions leads to global inconsistencies across segments and substantial semantic loss or redundancy during suboptimal summarization.
To tackle these challenges, we propose \textbf{GLaVE-Cap}, a \textbf{G}lobal-\textbf{L}ocal \textbf{a}ligned video caption framework with \textbf{V}ision \textbf{E}xpert integration, which provides highly fine-grained and consistent video caption across both local and global perspectives. Our GLaVE-Cap framework consists of two core modules: TrackFusion and CaptionBridge. \textbf{TrackFusion} aims to enhance the comprehensiveness of local captions while ensuring inter-frame consistency. It leverages a video-specific expert model~\cite{GD,SAM2} to track objects across frames, which can better guide VLMs and enrich local annotations.
Additionally, to alleviate overemphasis on inter-frame dynamics, it generates local captions via a dual-stream structure, where one stream processes inter-frame dynamics while the other extracts intra-frame static details, thereby effectively preserving the expert-refined information in local captions.
\textbf{CaptionBridge} aims to integrate local and global captions. It first generates an overview caption, which can provide contextual guidance to the local captioning process, thereby reducing ambiguity and inconsistency. In parallel, it performs adaptive scene segmentation and summarization based on the semantic content of local captions, thereby ensuring semantic coherence within the scene and preventing information loss and semantic redundancy due to suboptimal summarization.

For evaluating the quality of video captions, a common approach is to assess to what extent a caption can serve as a proxy for the video in answering content-related questions~\cite{auroracap, chen2025vidcapbench}. However, existing captioning benchmarks suffer from low annotation quality, consisting only of single-scene videos, and sparse question coverage, limiting their use to multi-scene video captioning task. Moreover, general-purpose VQA benchmarks~\cite{videomme,mvbench,xiao2021next} typically offer few questions per video, limiting their ability to thoroughly assess caption content.
To fill this gap, we construct a manually refined fine-grained Video-QA benchmark \textbf{GLaVE-Bench}, which includes multi-scene videos and 5$\times$ more queries per video than existing benchmarks~\cite{mvbench,videomme,xiao2021next,chen2025vidcapbench}, covering diverse visual aspects for a comprehensive evaluation of fine-grained video captioning.

We apply GLaVE-Cap on various cutting-edge VLM backbones such as GPT-4o~\cite{gpt4o} and Qwen2.5-VL-72B~\cite{bai2025qwen2}, both achieving state-of-the-art (SOTA) performance on widely used video question-answer (VQA) benchmarks such as Video-MME~\cite{videomme}, MVBench~\cite{mvbench}, and VidCapBench~\cite{chen2025vidcapbench}. These experimental results provide strong evidence for the high quality and generalizability of GLaVE-Cap. 
In addition, we leverage GLaVE-Cap to generate an extensive training dataset \textbf{GLaVE-1.2M}. This dataset includes 16K fine-grained video captions and 1.2M corresponding QA pairs. The GLaVE-1.2M dataset serves as a robust resource for advancing research on fine-grained video understanding capabilities of Video-LLMs. We further train a lightweight student model \textbf{GLaVE-7B} on the GLaVE-1.2M dataset and also achieve notable performance, demonstrating the effectiveness of our fine-grained supervision.

In a nutshell, we summarize our four main contributions as follows:
\vspace{-2.5mm}
\begin{itemize}
        \item We propose GLaVE-Cap, a global-local aligned video captioning framework with vision expert integration. It can provide highly fine-grained and consistent video captions across both local and global perspectives.
        \item We introduce two modules in GLaVE-Cap: TrackFusion leverages vision experts and a dual-stream design to enable frame-consistent tracking and generate comprehensive local captions; CaptionBridge aligns local captions and performs adaptive scene-level summarization via local-global interaction, ensuring fine-grained and consistent video captions.
        \item We present GLaVE-Bench, a comprehensive benchmark with 5$\times$ more queries per video than existing benchmarks, covering diverse visual dimensions for reliable evaluation of fine-grained video captioning. Additionally, we construct GLaVE-1.2M, a large-scale dataset with fine-grained QA pairs to advance fine-grained video understanding in Video-LLMs.
        \item We apply GLaVE-Cap to various VLM backbones, achieving SOTA performance on multiple pubic benchmarks and GLaVE-Bench. Additionally, we train a student model, GLaVE-7B, to explore the effectiveness of our fine-grained video captions and QAs.
\end{itemize}
\section{Related Work}
\textbf{Video-Language Models.}
Recent years have witnessed remarkable progress in video understanding through the development of large vision language models. Early explorations~\cite{DBLP:conf/emnlp/ZhangLB23,DBLP:conf/emnlp/LinYZCNJ024, DBLP:conf/acl/0001RKK24, li2024llama, zhang2024direct} established foundational approaches to align the video and text modalities.
Building upon these foundations, recent works~\cite{lin2024vila, li2024llava, llava_video,bai2025qwen2, chen2024expanding} further demonstrated that vanilla image language models that support multi-image as input can generalize to video understanding in a zero-shot manner and achieve improved performance when finetuned on video data. These works have helped standardize the practices for model architecture and training strategies.
Despite variations in design, these models commonly depend on large-scale, high-quality video datasets to improve video understanding capabilities. In this work, we focus on the task of video detailed captioning, which not only facilitates modality alignment training but also serves as a basis for generating complex supervision signals.

\textbf{Video Captioning Pipeline.}
High-quality video text annotations are crucial for modality alignment when training large-scale video-language models. Early approaches~\cite{miech2019howto100m, zellers2021merlot, xue2022advancing} primarily relied on metadata or ASR-generated subtitles to annotate videos at scale. With the rapid advancement of vision language models~\cite{li2023blip, zhu2023minigpt, DBLP:conf/emnlp/ZhangLB23}, some works~\cite{wang2023internvid, chen2024panda} have explored the use of these models to generate more semantically aligned video captions. Although these methods significantly improve annotation quality, they typically focus on short video clips, limiting their ability to handle longer and more complex videos and generate dense and fine-grained descriptions. To address this limitation, some recent studies~\cite{vript, auroracap} employ more capable image language models, such as GPT-4V, and introduce dedicated annotation pipelines, enhancing both the quality and temporal coverage of captions for longer videos. ~\cite{sharegpt4video, llava_video, peng2024inst} further refine the annotation pipeline to generate more detailed and temporally-grounded video captions, while scalable to larger number of video frames. 
In this paper, we observe that although existing methods propose reasonable pipelines and carefully craft text prompts to get accurate video annotations from VLMs, their effectiveness remains limited. VLMs often struggle to capture fine-grained details, maintain contextual coherence, and are prone to hallucinations. In contrast, our approach integrates multiple mechanisms to achieve the best detailed video annotation performance.

\section{Method~\label{sec:method}}
\begin{figure}[t]
  \centering
  \includegraphics[width=1.0\textwidth]{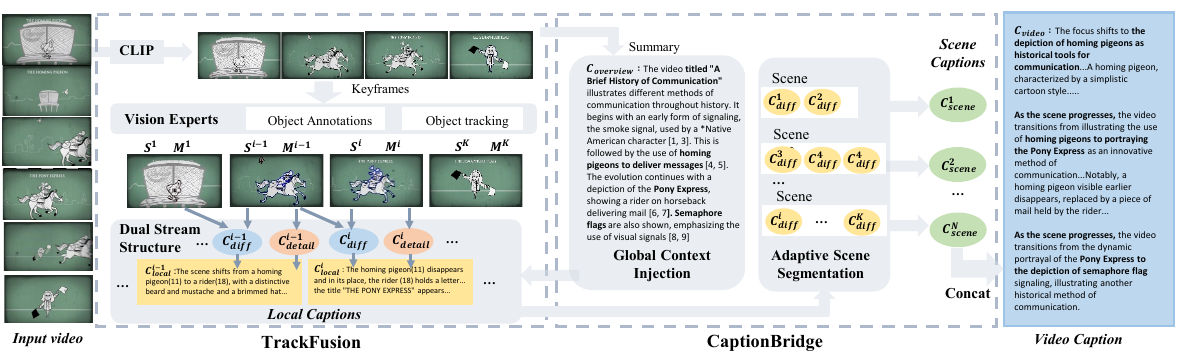}
  \caption{Overview of GLaVE-Cap. Given a video, TrackFusion first integrates vision experts to extract keyframes and track objects, then generates fine-grained local captions in a dual-stream manner. CaptionBridge injects the global context into local captions and achieves adaptive scene segmentation and summarization via their semantic information.}
  \label{fig:pipline}
  \vspace{-1.0em}
\end{figure}

In this section, we introduce GLaVE-Cap to generate highly fine-grained and consistent video captions across both local and global perspectives, as illustrated in \cref {fig:pipline}. We first present TrackFusion in \cref{sec:TrackFusion}, which utilizes vision experts and a dual-stream structure to produce fine-grained local captions. We then elaborate on CaptionBridge in \cref{sec:CaptionBridge}, which leverages the local-global interaction mechanism to generate coherent and detailed video descriptions.
\subsection{Pipeline Overview~\label{sec:pipeline_overview}}
The overall framework of GLaVE is illustrated in~\cref{fig:pipline}. Given an input video $V$, we first apply CLIP~\cite{CLIP} to extract keyframes $\mathcal{K}$. The extracted keyframes are then processed by TrackFusion module to obtain object masks by utilizing vision experts. The masked keyframes $\mathcal{M}$ and corresponding textual supplementary information $\mathcal{S}$ are integrated to generate two types of local captions: a differential caption $C_\texttt{diff}$ capturing dynamic changes and a detailed caption $C_\texttt{detail}$ focusing on static attributes. These are then integrated into a unified local caption $C_\texttt{local}$.
The CaptionBridge module first generates an overview caption $C_\texttt{overview}$ from the keyframes $\mathcal{K}$ and injects it into the local caption generation process. It then derives scene segmentation $\mathcal{SS}$ based on $C_\texttt{local}$ and $C_\texttt{overview}$ and summarizes each segment with the corresponding scene caption $C_\texttt{scene}$. These scene captions are finally concatenated to form the video caption $C_\texttt{video}$.

\subsection{TrackFusion\label{sec:TrackFusion}}
VLMs have demonstrated strong capabilities in capturing visual details~\cite{gpt4o,bai2025qwen2}. Leveraging this strength, most existing local captioning methods~\cite{sharegpt4video,vript,llava_video} directly feed multiple key-frames into the model with only prompt-level guidance to generate local captions in a single pass. However, VLMs still struggle with understanding object spatial relations and quantities~\cite{llava,otter,gpt4o}, which limits the completeness and accuracy of the resulting local captions. Furthermore, due to the sequential nature of the keyframe inputs, these single-pass approaches tend to emphasize dynamic visual changes while neglecting fine-grained static information, thereby failing to fully exploit the fine-grained perceptual capabilities. To address these limitations, we introduce TrackFusion, a framework that incorporates vision experts to perform cross-frame consistency tracking, enhancing the model’s ability to capture object counts and spatial relationships. In addition, we design a dual-stream architecture that separately encodes dynamic variations and static details, and integrates them to generate more comprehensive local captions.

\noindent\textbf{Track Objects Across Frames Via Vision Experts.} 
To address the limited capability of capturing and tracking objects in visual contents, recent works introduce image experts to highlight salient objects and generate visual tags~\cite{som,woodpecker}. However, extending this strategy to video is non-trivial due to frequent object movements, appearances, and disappearances over time. To overcome this challenge, we leverage Grounding DINO~\cite{GD} and SAM 2~\cite{SAM2} to achieve frame-level object consistency through tracking. Nevertheless, initializing bounding boxes with Grounding DINO on the first frame and then applying SAM 2’s video predictor for subsequent frames struggles to detect and track newly emerging objects. To mitigate this issue, inspired by grounded-SAM 2~\cite{grounded-sam2}, we integrate Grounding DINO and SAM 2's image predictor to extract bounding boxes $\mathcal{S}$ and object masks for each keyframe. Newly detected objects are assigned a new ID if they have insignificant overlap with existing tracked instances; otherwise, they inherit the ID of the most similar tracked object. The set of image masks with their IDs are then used to update the SAM 2 video predictor, ensuring stable and consistent object segmentation throughout the video. After extracting object masks, for each keyframe, we draw high-contrast boundaries and place numeric IDs at object centers as visual prompts following Set-of-Mark~\cite{som}, generating masked keyframes $\mathcal{M}$ for subsequent local caption generation.


\begin{figure}[t]
  \centering
  \includegraphics[width=1.0\textwidth]{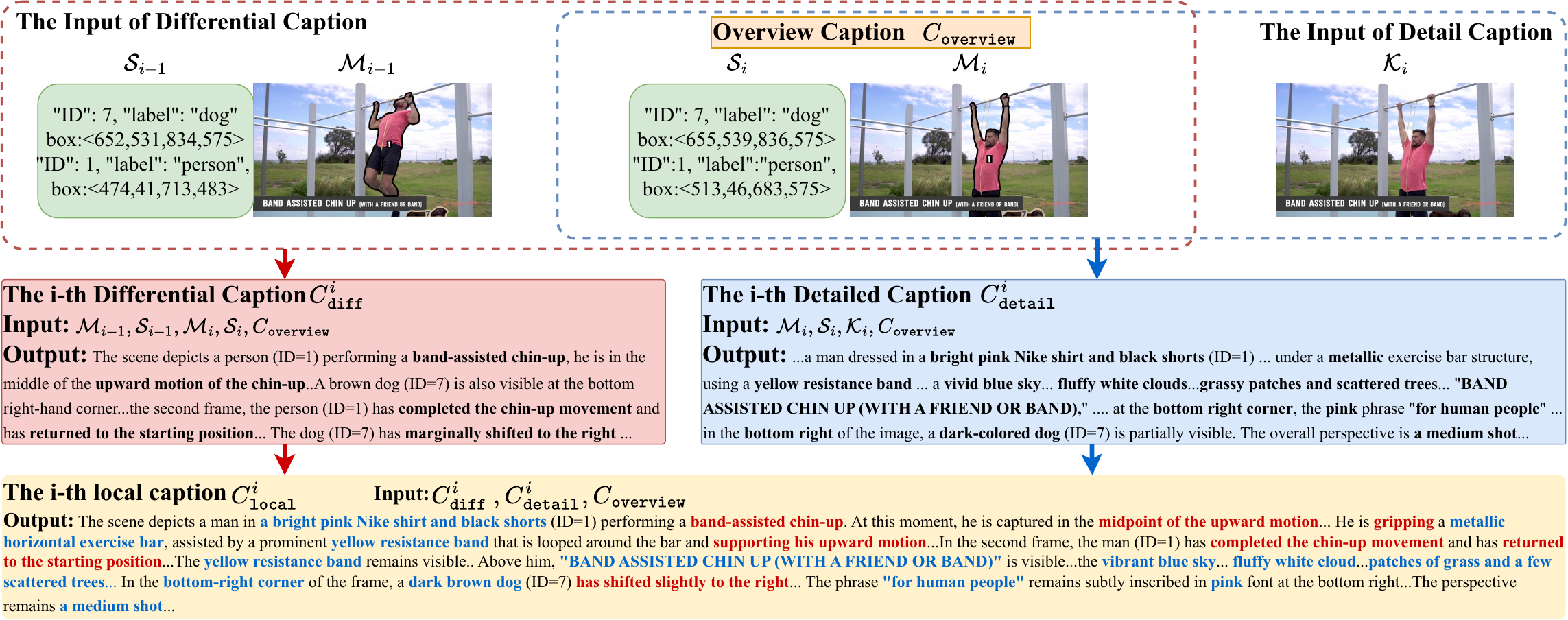}
  \caption{Illustration of the dual-stream structure introduced in TrackFusion.}
  \label{fig:local}
  \vspace{-1.0em}
\end{figure}
\noindent\textbf{Dual-Stream Structure.}
To overcome the dynamic bias in single-pass local caption generation method, we introduce a dual-stream local captioning structure that decouples the capturing of dynamic and static information, thereby maximizing the detail-capturing capacity of VLMs enhanced by vision experts. Specifically, we first generate a differential caption $C_\texttt{diff}$ to capture actions and changes across adjacent keyframes. Then, by comparing the original keyframes with their visually prompted counterparts, we produce a detailed caption $C_\texttt{detail}$ focusing on scene context and object attributes. To mitigate object reference ambiguity during merging, the VLM is instructed to append the unique ID of each annotated object when generating both $C_\texttt{diff}$ and $C_\texttt{detail}$. Finally, these two temporary captions are merged into a comprehensive local caption $C_\texttt{local}$. The generation process of local captions can be formulated as follows:
\begin{equation}
    \label{DFC}
    C_\texttt{diff}^i = \texttt{VLM}(\mathcal{M}_{i-1}, \mathcal{M}_i, \mathcal{S}_{i-1}, \mathcal{S}_i, C_\texttt{overview} \mid \texttt{prompt}_\texttt{diff})
\end{equation}
\begin{equation}
    \label{DTC}
    C_\texttt{detail}^i = \texttt{VLM}(\mathcal{K}_i, \mathcal{M}_i, \mathcal{S}_i, C_\texttt{overview} \mid \texttt{prompt}_\texttt{detail})
\end{equation}
\begin{equation}
    \label{MDC}
    C_\texttt{local}^i = \texttt{VLM}(C_\texttt{diff}^i, C_\texttt{detail}^i, C_\texttt{overview} \mid \texttt{prompt}_\texttt{merge})
\end{equation}
where $\mathcal{K}$ and $\mathcal{M}$ denote the original keyframes and their visually prompted counterparts. $C_\texttt{overview}$ denotes the overview caption used to inject the global information, which will be detailed below.

\subsection{CaptionBridge}\label{sec:CaptionBridge}
The existing local-to-global captioning frameworks~\cite{vript,sharegpt4video,llava_video} summarize the video caption from local captions. This limited interaction leads to two significant limitations. First, the absence of global context in local captions results in inconsistent descriptions of the same object across different video segments, thereby undermining the temporal coherence of the video caption. Second, directly summarizing inevitably omits a substantial amount of detailed information, while for the methods which involves scene splitting, methods relying solely on visual changes~\cite{vript,chen2024panda} may over-segment the video into unnecessary scenes under frequent viewpoint shift, resulting in redundant descriptions. As a consequence, these approaches often lead to suboptimal scene segmentation, resulting in either significant information loss or redundancy. To address these challenges, we propose CaptionBridge, a bidirectional interaction framework between local and global captions. By incorporating a global context into the local caption generation process, we enhance the consistency of local descriptions; additionally, by leveraging the semantic information of local captions, we ensure semantic coherence within the scene, thereby mitigating information loss caused by suboptimal scene segmentation and summarization.

\noindent\textbf{Global Context Injection.}
Current methods typically rely on a limited set of keyframes and the preceding frame-level caption for generating local captions, lacking access to global context and future information. This local-only perspective often leads to inaccurate object recognition and action understanding, resulting in inconsistent video caption, as illustrated in~\cref{fig:teaser} (B). To alleviate this issue, we introduce an overview caption $C_\texttt{overview}$ that summarizes the global context and inject it into the local captioning process to reduce recognition errors and improve local caption consistency. Specifically, we feed all the key-frames $\mathcal{K}$ into the VLM to generate an overview caption that includes key visual elements, event descriptions, timeline understanding, etc. Additionally, we require the VLM to annotate the corresponding range of keyframes for each sentence in the overview caption, facilitating the localization and understanding during the generation of different captions and summarization:
\begin{equation}\label{eq_overview}
    C_\texttt{overview} = \texttt{VLM}(\mathcal{K}_{[1:n]} \mid  \texttt{prompt}_{\texttt{overview}})
\end{equation}
where n denotes the number of keyframes and $\mathcal{K}_{[1:n]}$ denote all the original keyframes.

\noindent\textbf{Adaptive Scene-level Segmentation and Summarization.}
To address the limitation of inaccurate scene split and summarization mentioned above, we propose an adaptive scene-level summarization strategy that leverages the semantics of local captions to guide the video caption generation process.

Specifically, we first employ PySceneDetect~\cite{pyscenedetect} to initialize scene segmentation, which typically produces accurate boundaries but may result in redundant segments due to viewpoint shifts. We then prompt VLM to merge redundant scenes by leveraging the semantic information from all local captions $C_\texttt{local}^{[1,n]}$ and the global context provided by the overview caption $C_\texttt{overview}$. This produces semantically coherent scene partitions with precise boundaries, which serve as the foundation for generating scene-level captions $C_\texttt{scene}$. Finally, by directly concatenating the relatively self-contained scene-level captions in temporal order, we can naturally obtain a comprehensive and fine-grained video-level caption. The overall generation process can be formulated as follows:
\begin{equation}\label{SS}
    \mathcal{SS} = \texttt{VLM}(C_\texttt{local}^{[1,n]}, \texttt{PSD} \mid \texttt{prompt}_\texttt{SS})
\end{equation}
\begin{equation}\label{SC}
    C_\texttt{scene}^i = \texttt{VLM}(C_\texttt{local}^{[\mathcal{SS}_i.\texttt{ST}, \mathcal{SS}_i.\texttt{ED}]}, C_\texttt{scene}^{i-1}, C_\texttt{overview} \mid \texttt{prompt}_\texttt{SC})
\end{equation}
where \texttt{PSD} the scene split result of PySceneDetect, $n$ denotes the number of local captions, $\mathcal{SS}_i.\texttt{ST}$ and $\mathcal{SS}_i.\texttt{ED}$ denote the start and end keyframe numbers of scene $i$.
\section{Benchmark \& Dataset}
\subsection{GLaVE-Bench}
\label{sec:benchmark}
Evaluating the task of video detailed captioning requires a large number of high-quality videos, particularly those featuring complex and multi-scene, along with extensive QA annotations within one video. However, existing datasets either contain only a limited number of scenes~\cite{chen2024autoeval,liu2024tempcompass,mvbench, xiao2021next,chen2025vidcapbench}, making it difficult to assess the effectiveness of video captioning methods in complex scenarios, or include too few questions per video~\cite{chen2024autoeval,mvbench, xiao2021next,videomme}, limiting the comprehensiveness of the evaluation. To fill this gap, we propose GLaVE-Bench, which includes high-quality multi-scene videos and provides over 5$\times$ QA pairs per video (118.02 vs. 16.55) than existing benchmark to provide a comprehensive evaluation of video captioning according to~\cref{tab:bench}. Furthermore, we provide an additional scene hint during evaluation to reduce ambiguity caused by changes in object attributes throughout the video, which is common in multi-scene videos.

\noindent\textbf{Scene Hints.}
In complex multi-scene videos, a single question may correspond to different answers depending on the specific scene being referenced. Without sufficient scene disambiguation before asking the question, answers can become ambiguous, leading to unreliable evaluation results. To address this, we provide manually verified scene-hints for each scene, which uniquely identify the scene without revealing the answer to any questions. Please refer to Appendix~\cref{appendix:example_scene_hints} for an example illustrating how the absence of scene-hint leads to unreliable evaluation results.

\begin{table}[t]
\centering
\caption{The comparison of various benchmarks. The number of scenes is estimated by PySceneDetect. Scene-Hints indicates whether scene hints are provided to prevent ambiguity.}
\label{tab:bench} 
\resizebox{1.0\linewidth}{!}{
\begin{tabular}{lccccccc}
\toprule
\textbf{Benchmarks} & Videos & Video Length(s) & Scenes/Video & QAs & QA/Video & Annotation & Scene-Hints \\
\midrule
NExT-QA~\cite{xiao2021next} & 1,000 & 39.5 & 2.95 & 8,564 & 8.56 & Auto & \XSolidBrush \\
MVBench~\cite{mvbench} & 3,851 & 16.0 & 1.43 & 4,000 & 1.04 & Auto & \XSolidBrush \\ 
AutoEval-Video~\cite{chen2024autoeval} & 327 & 14.6 & 3.37 & 327 & 1.00 & Manual & \XSolidBrush \\
TempCompass~\cite{liu2024tempcompass} & 410 & 11.4 & 1.16 & 7,540 & 15.87 & Auto\&Manual & \XSolidBrush \\ 
Video-MME-S~\cite{videomme} & 300 & 80.7 & 17.67 & 900 & 3.00 & Manual & \XSolidBrush \\ 
VidCapBench~\cite{chen2025vidcapbench} & 643 & 10.2 & 2.38 & 10,644 & 16.55 & Auto\&Manual & \XSolidBrush \\ 
GLaVE-Bench~\cite{mvbench} & 55 & 85.5 & 8.84 & 6,491 & \textbf{118.02} & Auto\&Manual & \Checkmark \\ 
\bottomrule
\end{tabular}
}
\vspace{-1.5em}
\end{table}
\noindent\textbf{Data Sources and Automatic Annotation.}~\label{sec:Annotation}
We adopt the Video-MME Short subset as the data source, as it offers high-resolution videos with clear thematic focus and diverse scenes. Then, we manually select 55 videos that feature multiple scenes, stable camera motion, and well-defined topics. Based on GLaVE-Cap, we first generate fine-grained video captions and summary scene-level hints, which is a concise sentence that accurately describes scene, to identify each scene. Then, we construct question-option pairs, where each question is accompanied by four candidate answers, only one of which is correct. Finally, we incorporate human verification to ensure the accuracy and overall quality of the data, which takes approximately 230 human annotation hours in total. For more question-option pairs generation and human review details, please refer to~\cref{appendix:GLaVE-Bench}.

\noindent\textbf{Evaluation Protocol}~\label{sec:eval}
Given a video caption and the corresponding question-option pair, we prompt GPT-4o~\cite{gpt4o} to choose the option from four with the scene hint to prevent ambiguity. We require GPT-4o to select option "E. Not mentioned" if the caption lacks sufficient information for answering the question. We report metrics to comprehensively assess the captioning model: $Acc.=\frac{n_c}{n_c+n_w+n_e}$, $Hall.=\frac{n_w}{n_c+n_w}$, and $N.M.=\frac{n_e}{n_c+n_w+n_e}$, where $n_c$, $n_w$, and $n_e$ denote the numbers of correct, non-E errors, and E-type responses, respectively. High $N.M.$ indicates the insufficient comprehensiveness of the caption, while high $Hall.$ reflects the hallucination. For scene-level questions, scene hints from adjacent scenes are additionally provided to reduce ambiguity.  

\subsection{GLaVE-1.2M}
With the fine-grained video captioning capability of GLaVE-Cap, we further construct a large-scale training dataset GLaVE-1.2M. The video source is LLaVA-Video-178K~\cite{llava_video}, which comprises 178K high-resolution, untrimmed videos from ten diverse sources, including HD-VILA-100M~\cite{xue2022advancing}, Activity-Net~\cite{activitynet}, and NeXT-QA~\cite{xiao2021next}.
We filter videos based on duration (30–180 seconds) first and then exclude those with too many or too few (less 2 or over 10) scenes using PySceneDetect. Finally, we assess the videos using GPT-4o~\cite{gpt4o} to ensure stable camera motion, clear thematic focus, and no frequent scene transitions, thus guaranteeing the selection of high-quality video content.

After filtering, we obtain our dataset consisting of 15,814 videos, totaling 376 hours. Using the GLaVE-Cap and QA generation method described in \cref{sec:Annotation}, we generate a total of 15,814 video captions and 1,176,410 QA pairs. For data distributions of GLaVE-1.2M please refer to \cref{appendix:GLaVE-1.2M}.
\section{Experiments}
\begin{table*}[!t]
  \caption{Quantitative comparison of video captioning strategies. The best results are \textbf{bold} and the second-best results are \underline{underlined}.}
  \centering
  \resizebox{0.99\linewidth}{!}{
  \begin{tabular}{llcccccccccccc}     
        \toprule
        \multicolumn{1}{l}{\multirow{2}{*}{Method}} & \multicolumn{1}{l}{\multirow{2}{*}{V-L backbone}} & \multicolumn{3}{c}{GLaVE-Bench} & \multicolumn{3}{c}{VidCapBench~\cite{chen2025vidcapbench}} & MME-S~\cite{videomme} & AS~\cite{mvbench} & OE~\cite{mvbench} & AC~\cite{mvbench} & CO~\cite{mvbench} & MD~\cite{mvbench} \\
        \cmidrule{3-14}
        \multicolumn{1}{l}{} & \multicolumn{1}{l}{} & $Acc.$ & $Hall.$$\downarrow$ & $N.M.$$\downarrow$ & $Acc.$ & $Pre.$ & $Cov.$ & $Acc.$ & $Acc.$ & $Acc.$ & $Acc.$ & $Acc.$ & $Acc$.  \\
        \midrule
        \multicolumn{1}{l}{\multirow{2}{*}{LVD-2M~\cite{lvd2m}}} & Qwen2.5-VL & 42.21 & 19.69 & 47.45 & 13.50 & 51.07 & 83.86 & 64.00 & 53.67 & 48.50 & 38.16 & 73.50 & 31.33 \\
        \multicolumn{1}{l}{} & GPT4o & 37.65 & 19.63 & 53.16 & 12.41 & 49.22 & 81.57 & 59.67 & 55.50 & 47.50 & 37.83 & 69.33 & 25.17 \\
        \midrule
        \multicolumn{1}{l}{\multirow{2}{*}{AuroraCap~\cite{auroracap}}} & Qwen2.5-VL & 40.00 & 20.93 & 49.41 & 15.94 & 56.08 & 86.60 & 58.93 & 62.83 & 66.16 & 38.33 & 70.83 & 43.67 \\
        \multicolumn{1}{l}{} & GPT4o & 41.16 & 21.71 & 47.42 & 16.01 & 55.84 & 86.34 & 59.79 & 67.50 & 59.17 & 37.67 & 71.83 & 34.67 \\
        \midrule
        \multicolumn{1}{l}{\multirow{2}{*}{ShareGPT4Video~\cite{sharegpt4video}}} & Qwen2.5-VL & 47.97 & 19.63 & 40.32 & 16.43 & 56.26 & 88.44 & 65.00 & 57.12 & 65.67 & 39.16 & 73.67 & 42.67 \\
        \multicolumn{1}{l}{} & GPT4o & 45.64 & 17.85 & 44.45 & 16.71 & 56.70 & 87.09 & 63.19 & 62.83 & 63.50 & 38.17 & 79.67 & 44.83 \\
        \midrule
        \multicolumn{1}{l}{\multirow{2}{*}{LLaVA-Video~\cite{llava_video}}} & Qwen2.5-VL & 46.56 & 19.65 & 42.06 & 15.47 & 54.49 & 83.67 & 70.33 & 64.83 & \underline{66.67} & 40.33 & 76.67 & 47.17 \\
        \multicolumn{1}{l}{} & GPT4o & 47.88 & 19.02 & 40.88 & 18.25 & 57.55 & 89.07 & 69.37 & \underline{69.00} & 65.00 & \underline{40.83} & \underline{80.17} & 39.33 \\
        \midrule
        \multicolumn{1}{l}{\multirow{2}{*}{Vript~\cite{vript}}} & Qwen2.5-VL & 50.94 & 19.02 & 37.10 & 14.91 & 56.53 & 85.80 & 69.59 & 56.83 & 59.00 & 34.33 & 67.83 & 45.00 \\
        \multicolumn{1}{l}{} & GPT4o & 55.31 & 18.58 & 32.06 & 17.62 & 57.92 & 87.93 & \underline{73.48} & 58.83 & 55.50 & 35.67 & 75.67 & 38.50 \\
        \midrule
        \midrule
        \multicolumn{1}{l}{\multirow{2}{*}{GLaVE-Cap}} & Qwen2.5-VL & \underline{63.67} & \underline{17.42} & \underline{22.90} & \underline{18.78} & \textbf{60.98} & \textbf{92.34} & 73.19 & 65.50 & \textbf{67.17} & 40.33 & 75.83 & \underline{55.50} \\
        \multicolumn{1}{l}{} & GPT4o & \textbf{67.68} & \textbf{16.04} & \textbf{19.40} & \textbf{19.40} & \underline{58.61} & \underline{91.79} & \textbf{74.52} & \textbf{70.67} & 64.33 & \textbf{42.17} &\textbf{81.17} & \textbf{59.33} \\
        \bottomrule
  \end{tabular}
  }
  \label{tab-main-caption}
  \vspace{-1.5em}
\end{table*}
\subsection{Video-Captioning Strategy Comparison~\label{sec:caption_result}}
\noindent\textbf{Compared methods.}
We select LVD-2M~\cite{lvd2m}, AuroraCap~\cite{auroracap}, ShareGPTVideo~\cite{sharegpt4video}, LLaVA-Video~\cite{llava_video}, and Vript~\cite{vript} for comprehensive comparison. Among them, ShareGPTVideo and LLaVA-Video adopt innovative approaches: DiffSW and multi-level strategies, which are effective in capturing both fine-grained details and global context. Vript employs a scene-wise summarization strategy, demonstrating stronger temporal consistency. For more implementation details of baseline methods, please refer to ~\cref{sec:reproduction-detail}

\noindent\textbf{Benchmarks \& Evaluation.}
We evaluate video-captioning strategies on multiple benchmarks, including GLaVE-Bench, VidCapBench~\cite{chen2025vidcapbench}, Video-MME-S~\cite{videomme}, and five subsets of MVBench~\cite{mvbench}: Action Sequence, Object Existence, Action Count, Character Order, and Moving Direction. 
GLaVE-Bench and VidCapBench primarily focus on the granularity of caption descriptions, with GLaVE-Bench being more challenging due to the multi-scene nature of the videos. Video-MME emphasizes the holistic comprehension of multi-scene video content, while MVBench provides a comprehensive evaluation using short, specialized video clips, with a particular focus on visual changes.

Following VidCapBench~\cite{chen2025vidcapbench}, we evaluate video captions by replacing the original videos with the generated captions and prompting GPT-4o~\cite{gpt4o} to answer or select an answer.
For GLaVE-7B, we report $Acc.$, $Hall.$, and $N.M.$ following the evaluation protocol described in ~\cref{sec:eval}. For VidCapBench, we report $Acc.$, $Pre.$, and $Cov.$. For the remaining VQA benchmarks, we report $Acc.$. We conducted three evaluations for each question to reduce the variability in GPT-based assessments.

\noindent\textbf{Experimental Results.}
\cref{tab-main-caption} presents a comparative evaluation of various video captioning strategies. LVD-2M, AuroraCap, and ShareGPT4Video lack well-designed summarization mechanisms, resulting in subpar performance across all four benchmarks. Though Vript performs comparably to our approach on Video-MME owing to its scene-wise summarization strategy, its intra-scene summarization is not comprehensive, resulting in inferior performance on other benchmarks. LLaVA-Video, with its multi-level caption strategy, demonstrates strong capability in capturing detailed visual information and changes within short videos in VidCapBench and MVBench. However, this approach tends to excessively compress intra-scene details in multi-scene videos, leading to suboptimal performance on GLaVE-Bench and Video-MME. In contrast to the aforementioned methods, GLaVE-Cap achieves SOTA results across all four benchmarks. These results demonstrate that, by introducing CaptionBridge, which includes vision experts and a dual-stream architecture for fine-grained local caption generation, our method can effectively handle tasks such as short video fine-grained description (VidCapBench) and change capture (MVBench). In parallel, the local-global interaction mechanism within CaptionBridge enables consistent and fine-grained video-level captions for complex multi-scene videos, supporting comprehensive video understanding and reasoning (Video-MME). When the benchmark focuses on fine-grained descriptions of multi-scene videos, both of our modules contribute effectively, leading to a significant performance advantage over other methods on GLaVE-Bench.
Furthermore, we achieve comparable performance when replacing the VLM with the open-source Qwen2.5-VL-72B~\cite{bai2025qwen2}. This demonstrates the generalizability of our video-captioning strategy. For more aspects of evaluation analysis, please refer to \cref{appendix:Analysis}.

\subsection{Student Model~\label{sec:model_result}}
\begin{table*}[!t]
  \caption{Quantitative comparison of video language models. * denotes our reproduction result.}
  \centering
  \resizebox{0.99\linewidth}{!}{
    \begin{tabular}{lcccc}
      \toprule
      \multirow{2}{*}{Models} & \multicolumn{1}{c}{GLaVE-Bench} & \multicolumn{2}{c}{Video-MME-S~\cite{videomme}} & \multicolumn{1}{c}{MV-Bench~\cite{mvbench}} \\
      \cmidrule{2-5}
      & $Acc.$  & w/o-sub $Acc.$ & w-sub $Acc.$ & $Acc.$ \\
      \midrule
      ShareGPT4Video-8B~\cite{sharegpt4video} & 51.52 & 39.9 & 43.6 & 51.2 \\
      LLaVA-OneVision-7B~\cite{li2024llava} & 72.98 & 58.2 & 61.5 & 56.7 \\
      LLaVA-Video-7B~\cite{llava_video} & 76.27 & 63.3 & 69.7 & 58.6  \\
      \midrule
      Qwen2.5-VL-7B$^{*}$ & 76.31 & 64.26 & 69.52 & 66.80 \\
      \midrule
      Qwen2.5-VL-7B-Vript & 75.50 & 61.04 & 65.56 & 66.33 \\
      Qwen2.5-VL-7B-ShareGPTVideo & 76.35 & 62.67 & 66.74 & 65.85 \\
      Qwen2.5-VL-7B-LLaVA-Video & 75.92 & 60.19 & 64.48 & 67.47 \\
      GLaVE-7B & 79.28 & 62.96 & 67.78 & 66.33 \\
      \bottomrule
    \end{tabular}
    }
  \label{tab-models}
\end{table*}

To evaluate the fine-grained characteristics of our GLaVE-1.2M dataset and its effectiveness in enhancing a model's fine-grained understanding capacity, we train a lightweight model, GLaVE-7B, using Qwen2.5-VL-7B~\cite{bai2025qwen2} as the base model. We then compare GLaVE-7B with the SOTA models~\cite{sharegpt4video,li2024llava,llava_video}, on GLaVE-Bench and other widely used VQA benchmarks~\cite{videomme,mvbench}. For a fair comparison, we also train three models based on the Vript~\cite{vript}, ShareGPTVideo~\cite{zhang2025direct}, and LLaVA-Video~\cite{llava_video} datasets using the same base model and training setup, named Qwen2.5-VL-7B-Vript, Qwen2.5-VL-7B-ShareGPTVideo, and Qwen2.5-VL-7B-LLaVA-Video, respectively. the composition of other datasets used to train Qwen2.5-VL-7B, please refer to \cref{appendix:Training-details}

\cref{tab-models} shows that GLaVE-7B achieves the highest performance on GLaVE-Bench, surpassing both prior SOTA models and Qwen2.5-VL-7B fine-tuned on other datasets. On Video-MME and MVBench, GLaVE-7B exhibits a slight performance decline compared to the base model, which we attribute to catastrophic forgetting~\cite{zhai2023investigating,li2024revisiting}, a phenomenon also observed in other Qwen2.5-VL-7B fine-tuned models. Notably, the performance decline of GLaVE-7B is consistently smaller than that of other models, indicating that GLaVE-1.2M provides more detailed and comprehensive supervision than other open-source datasets.

\begin{table*}[t]
    \centering
    \begin{minipage}[t]{0.525\textwidth}
        \centering
        \caption{Ablation studies on GLaVE-Bench. The VLM is GPT-4o~\cite{gpt4o}}
        \label{tab-ablation}
        \resizebox{\linewidth}{!}{
        \begin{tabular}{c@{\hspace{4pt}}c@{\hspace{4pt}}c@{\hspace{4pt}}c@{\hspace{4pt}}c@{\hspace{4pt}}c@{\hspace{4pt}}c}     
            \toprule
            \makecell{Visual\\Prompt} & \makecell{Overview\\Caption} & \makecell{Dual-stream\\Structure} & \makecell{Adaptive\\Scene-split} & Acc. & Hall.$\downarrow$ & N.M.$\downarrow$ \\
            \midrule
            \Checkmark & \Checkmark & \Checkmark & \XSolidBrush & 45.98 & 18.53 & 43.38 \\
            \Checkmark & \Checkmark & \XSolidBrush & \Checkmark & 62.67 & 17.31 & 24.21 \\
            \Checkmark & \XSolidBrush & \Checkmark & \Checkmark & 66.77 & 16.63 & 19.91 \\
            \XSolidBrush & \Checkmark & \Checkmark & \Checkmark & 67.09 & 15.90 & 20.22 \\
            \Checkmark & \Checkmark & \Checkmark & \Checkmark & 67.68 & 16.04 & 19.40 \\
            \bottomrule
        \end{tabular}
        }
    \end{minipage}%
    \hfill
    \begin{minipage}[t]{0.45\textwidth}
        \centering
        \caption{Ablation of visual prompt. VLM: GPT-4o~\cite{gpt4o}. Metric: $Acc$.}
        \label{tab-ablation-mask}
        \vspace{2pt}
        \resizebox{\linewidth}{!}{

        \begin{tabular}{lcc}     
            \toprule
            \makecell{Dataset / Visual Prompt} & \XSolidBrush & \Checkmark \\
            \midrule
            GLaVE-Bench (Spatial) & 48.13 & 53.14 \\
            GLaVE-Bench (Count) & 57.19 & 58.63 \\
            GLaVE-Bench (Direction) & 41.89 & 45.42 \\
            MVBench (Number count) & 44.17 & 51.33 \\
            NExT-QA (Moving count) & 58.07 & 62.94 \\
            \bottomrule
        \end{tabular}
        }
    \end{minipage}
    \vspace{-1.0em}
\end{table*}


\subsection{Ablation Studies~\label{sec:ablation}}
\cref{tab-ablation,tab-ablation-mask} present the results of our ablation studies. For the detailed experimental settings, please refer to \cref{appendix:Ablation}.
Removing the dual-stream structure and the adaptive scene split module leads to a notable decrease in accuracy (5.01\% and 21.70\%, respectively), indicating that these components play a crucial role in enhancing the granularity of in-scene descriptions and mitigating information loss during summarization, respectively. Although ablating the overview caption module does not result in a substantial drop in overall performance, it severely compromises the temporal coherence of video captions, thereby implicitly affecting annotation quality, which is not easy to evaluate, as shown in \cref{fig:teaser} (b). Although the removal of visual prompts does not significantly impact performance on GLaVE-Bench,  \cref{tab-ablation-mask} demonstrates that visual prompt ablation causes marked performance degradation in sub-tasks related to object counting across different datasets, suggesting the importance of the integration of vision experts for object quantity perception.

To provide a more comprehensive and accurate understanding of how different components of GLaVE-Cap affect performance, we additionally include extended experiments and analyses in the appendix. Specifically, we present direct ablations on each module, investigate the influence of prompts and model capabilities, and report more ablation results on other benchmarks. Please refer to \cref{appendix:direct_ablation,appendix:prompt_model,appendix:more_ablation} for details, respectively.
\section{Conclusion}
In this paper, we propose GLaVE-Cap, a novel framework for video detailed captioning that addresses the limitations of existing local-to-global paradigms through vision expert integration and local-global interaction.
GLaVE-Cap comprises two core modules: TrackFusion, which leverages vision experts and the dual-stream architecture to produce fine-grained, object-aware local captions; and CaptionBridge, which incorporates global context and performs adaptive scene-level summarization to generate coherent and detailed video-level captions.
To enable more comprehensive evaluation of video understanding, we introduce GLaVE-Bench, a benchmark that spans diverse visual dimensions. We also release GLaVE-1.2M, a large-scale dataset with high-quality captions and QA pairs to support training and analysis.
Extensive experiments demonstrate that GLaVE-Cap achieves state-of-the-art performance across various benchmarks. Further ablation studies and student model analyses validate the effectiveness of each proposed module and highlight the practical utility of GLaVE-1.2M.

{\small
\bibliographystyle{plain}
\bibliography{main}
}
\clearpage
\appendix
\section{Appendix Overview}
In the appendix, we provide more results and showcases. They are structured as follows:

\begin{itemize}
    \item More detailed Information of GLaVE-Bench in \cref{appendix:GLaVE-Bench}.
    \item Visualization of data distributions for GLaVE-1.2M in \cref{appendix:GLaVE-1.2M}.
    \item Implementation details of baseline captioning methods in \cref{sec:reproduction-detail}.
    \item Additional evaluation analysis in \cref{appendix:Analysis}.
    \item Training Details of GLaVE-7B in \cref{appendix:Training-details}.
    \item Ablation Details of GLaVE-Cap in \cref{appendix:Ablation}.
    \item More evidence on proposed modules in GLaVE-Cap can address the claimed three issues existing in current methods in \cref{appendix:direct_ablation}.
    \item Analysis of how different types of prompts and capabilities of (M)LLMs affect model performance in \cref{appendix:prompt_model}.
    \item More ablation studies on other commonly used benchmarks in \cref{appendix:more_ablation}.
    \item Hallucination Accumulation Analysis in \cref{appendix:Hallucination}.
    \item A caption generated by GLaVE-Cap is shown in \cref{appendix:caption_show}.
    \item Prompt templates used in GLaVE-Cap for caption generation in \cref{appendix:frame_prompt}.
    \item Prompt templates used in GLaVE-Bench for evaluation in \cref{appendix:eval_prompt}.
    \item Discussion of limitations and societal impacts in \cref{appendix:limitation}.
    \item License information in \cref{appendix:license}.
\end{itemize}

\section{More Information of GLaVE-Bench}\label{appendix:GLaVE-Bench}
\subsection{Question-Option Pairs Generation}
After generating video captions using GLaVE-Cap, we first construct question-answer pairs. Following the 16 question types defined in LLaVA-Video~\cite{llava_video}, we categorize them into two groups: scene-level QA and global-level QA. The scene-level QA targets fine-grained understanding within individual scenes, whereas the global-level QA assesses holistic comprehension and reasoning across the entire video.

For scene-level QA, we focus on 12 description-oriented question types, such as description-Object, fine-grained action, and count. For each scene $i$, we generate at most one question–answer pair per type based on the corresponding scene caption $C_\texttt{scene}^i$, ensuring broad coverage of nearly all scenes and their specific content. To reduce ambiguity, we adopt the scene hint strategy introduced in VRIPT-RR~\cite{vript}, generating a scene hint for each scene and prepending it to the corresponding question. In addition, we introduce a new question type, Visual-cue, which targets subtle yet informative visual elements—such as text, logos, and other cues—that are crucial for accurately understanding the video content.

For global-level QA, which comprises the remaining four question types targeting temporal order and causal reasoning, we generate up to five question–answer pairs per type based on the full video caption $C_\texttt{video}$. As these question types demand an understanding of long-range dependencies across the video, they can only be effectively constructed from the global context provided $C_\texttt{video}$.

After generating the initial question–answer pairs, we observed that some were not atomic, \ie, they could be decomposed into multiple independent QA pairs. Additionally, since the QA pairs were derived directly from captions, the questions and answers often exhibited strong continuity, with many questions implicitly revealing or suggesting their corresponding answers. To mitigate these issues, we employed GPT-4o~\cite{gpt4o} to refine the QA pairs by splitting non-atomic questions and rephrasing them to ensure greater neutrality and reduced answer leakage.

Finally, we generate the question–option pairs. In addition to the refined QA pairs, we provide GPT-4o~\cite{gpt4o} with the corresponding scene or video caption as contextual guidance. Based on this input, GPT-4o is asked to produce four answer options for each question: one correct answer and three distractors. The distractors are designed to be stylistically similar to the correct answer but factually incorrect, ensuring the plausibility of all options while maintaining a single ground-truth answer.

\begin{figure}[t]
  \centering
  \includegraphics[width=1\textwidth]{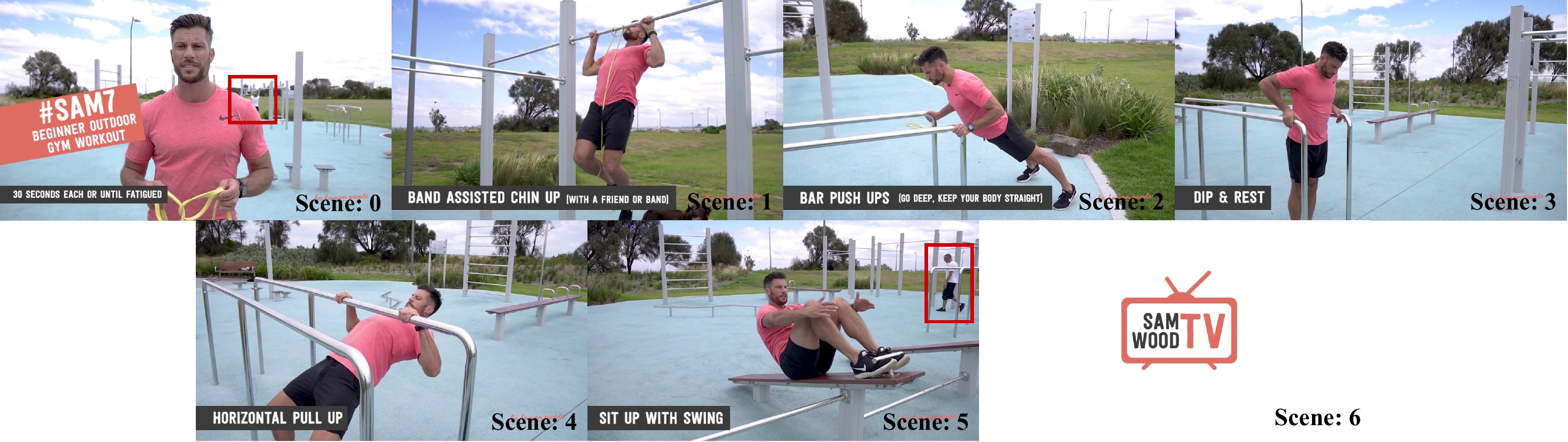}
  \caption{Illustration of the example video \texttt{6EIrArTyLVU.mp4} in GLaVE-Bench.}
  \label{fig:scene_hint}
\end{figure}

\subsection{Examples of Scene-Hints in GLaVE-Bench}\label{appendix:example_scene_hints}
For instance, to illustrate how the absence of scene-hint can lead to unreliable evaluation results, consider the video \texttt{6EIrArTyLVU.mp4} in our benchmark, which is shown in~\cref{fig:scene_hint}. Without any scene hint, the 55-th question: \textit{``In the early stage of the scene, compared to the man who is exercising, where is the person in the background located?''} could be interpreted differently depending on the referenced scene.

\begin{itemize}
  \item If \textbf{scene 0} is considered, the answer is: \textit{``The person is directly behind the man, using gym equipment in the background.''}
  \item If \textbf{scene 5} is considered, the answer becomes: \textit{``The person is located on the left side of the scene, walking across the rubber floor.''}
  \item For the rest of the scenes, the answer should be: \textit{``The man who is exercising is the only person.''}
\end{itemize}

To address this, we provide manually verified scene-hints for each scene, which uniquely identify the scene without revealing the answer to any downstream questions. These hints help eliminate ambiguity while preserving fairness and answer neutrality. For example, the scene-hints for this video are:

\begin{itemize}
  \item 0: A man is introducing something in front of the camera.
  \item 1: A man is doing pull-ups.
  \item 2: A man is doing dips on a bar.
  \item 3: A man is doing tricep dips.
  \item 4: A man is doing horizontal pulls.
  \item 5: A man is doing sit-ups.
  \item 6: A sign is displayed in the center.
\end{itemize}

These scene-hints ensure reliable evaluation by anchoring the question to the intended context, without leaking potential answers.

\subsection{Human Review}
After the initial automatic generation, we incorporate human verification to ensure the accuracy and overall quality of the data. Human annotators are instructed to carry out two main types of validation tasks. First, they verify the scene segmentation and associated hints, checking whether the segmentation is reasonable, whether the scene hints clearly refer to the intended scenes, and whether any unnecessary scene details are revealed. Revisions are made when segmentation or hints are found to be inappropriate. Second, annotators evaluate each question–option pair to verify the correctness of the question–answer pairs, eliminate ambiguity, and ensure consistency between the scene-level QA pairs and the referenced scene. When issues are identified, the QA pairs are either revised or removed. Note that the tasks mentioned above represent typical examples of the verification process; the full annotation guidelines are provided in the supplementary material: Annotation.md. After human verification, a total of 6,491 question–option pairs are retained, with a total of 230 rater-hours spent.
\begin{figure}[t]
  \centering
  \includegraphics[width=1.0\textwidth]{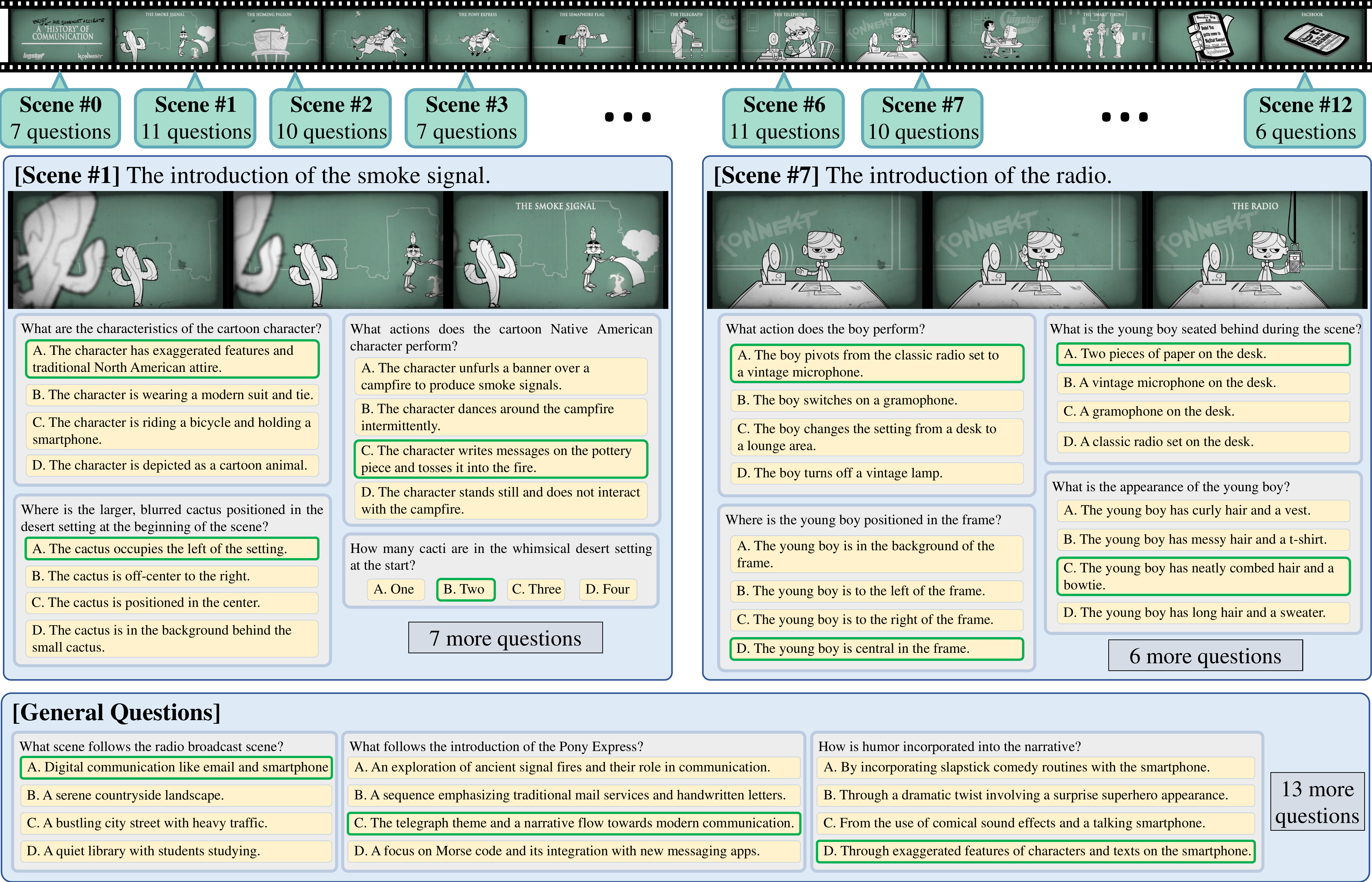}
  \caption{A case of GLaVE-Bench. The video contains 13 scenes and 109 scene questions in total.}
  \label{fig:benchmark_case}
\end{figure}
\subsection{Showcase}
Please refer to \cref {fig:benchmark_case} for the showcase.

\begin{figure}[t]
  \centering
  \includegraphics[width=1.0\textwidth]{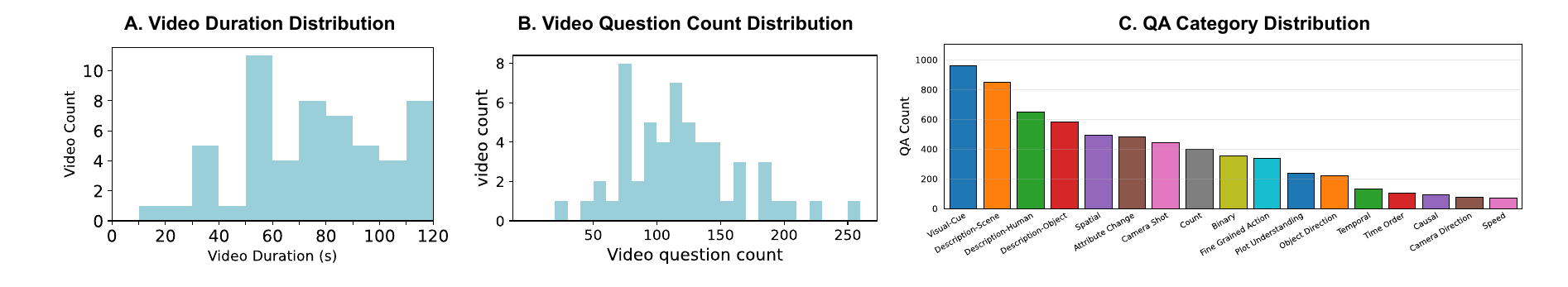}
  \caption{Visualization of various distributions in the test benchmark.}
  \label{fig:benchmark}
\end{figure}
\subsection{Visualization of data distributions}
Please refer to \cref{fig:benchmark} for the distribution of video duration (A), video question count (B), and QA category (C).

\begin{figure}[t]
  \centering
  \includegraphics[width=1.0\textwidth]{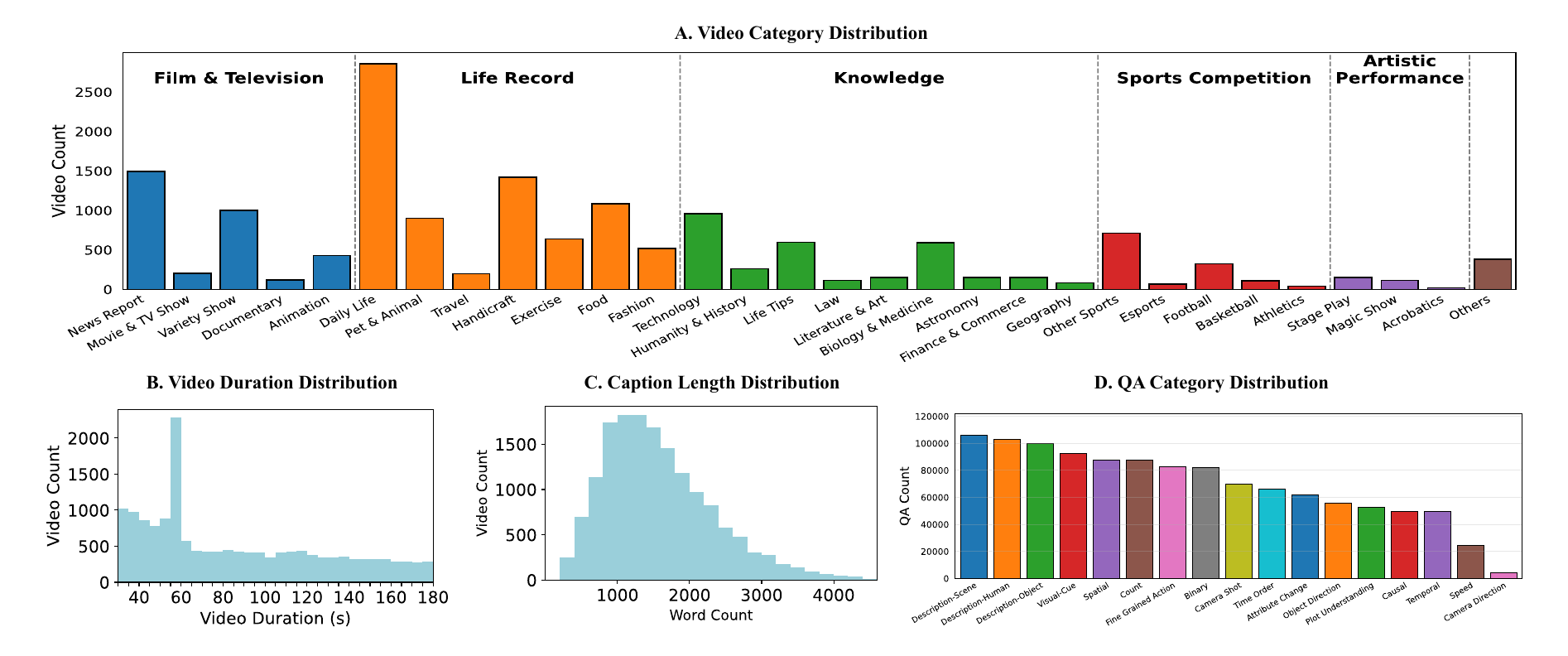}
  \caption{Visualization of various distributions in the training dataset.}
  \label{fig:dataset}
\end{figure}
\section{Visualization of Data Distributions in GLaVE-1.2M}\label{appendix:GLaVE-1.2M}
Please refer to \cref{fig:dataset} for the distribution of video category (A), video duration (B), caption length (C), and QA category (D).

\section{Implementation Details of Baseline Captioning Methods}
\label{sec:reproduction-detail}
We selected several representative video captioning pipelines proposed in previous works. As many of these works did not release their complete pipeline code, we implemented the methods based on the descriptions provided in their respective papers. In our reproductions, we preserved the original prompt templates, frame sampling strategies, and multi-step annotation procedures whenever available. However, for a fair comparison, we standardized certain components across all methods, such as the choice of annotation model and the input image resolution.

In the following, we outline the implementation details of each baseline pipeline used in our experiments.

\begin{itemize}
\item \textbf{LVD-2M}~\cite{lvd2m}: This method generates a detailed caption for every 30-second video segment by sampling 6 frames as input to the annotation model, followed by refinement and summarization. In our implementation, the 6 frames are provided as separate high-resolution images, rather than as a single $2 \times 3$ grid image as originally reported.

\item \textbf{AuroraCap}~\cite{auroracap}: This method first produces a structured caption and then refines it into a detailed caption. Since the original paper did not specify a frame sampling strategy, we chose to sample at 1 frame per second, or use a fixed 32-frame sampling strategy for videos longer than 32 seconds.

\item \textbf{ShareGPT4Video}~\cite{sharegpt4video}: This work proposes the DiffSW captioning pipeline, which we implemented faithfully according to the description provided in the paper.

\item \textbf{LLaVA-Video}~\cite{llava_video}: This method introduces a three-stage aggregative captioning pipeline. We reproduced the pipeline using the official code released by the authors.

\item \textbf{Vript}~\cite{vript}: This method divides the video into multiple clips and generates captions for each clip individually. The original implementation includes audio transcripts as input during caption generation, but we excluded it in our reproduction as audio is beyond the scope of this work.
\end{itemize}

\section{Additional Evaluation Analysis}
\label{appendix:Analysis}
\subsection{Sample Frame Count Comparison}
As discussed in \cref{sec:reproduction-detail}, we did not unify the video frame sampling strategy between captioning methods, since some methods lack support for flexible frame sampling and the effect of frame count on caption quality remains uncertain. To improve the credibility of our comparison, we report the average number of frames used for caption generation in \cref{tab:fps-comparison}. Some methods process more input frames, theoretically accessing more video information, but still fail to produce better captions.

\begin{table}[htbp]
\centering
\caption{Average sample frame count comparison across multiple methods and benchmarks}
\label{tab:fps-comparison} 
\vspace{1em}
\resizebox{0.99\linewidth}{!}{
\begin{tabular}{lcccccccc}
\toprule
\textbf{Methods} & \textbf{GLaVE-Bench} & \textbf{MME} & \textbf{AS} & \textbf{OE} & \textbf{AC} & \textbf{CO} & \textbf{MD} & \textbf{VidCap} \\
\midrule
LVD-2M~\cite{lvd2m}          & 15.80 & 16.15 & 6.11  & 1.00 & 4.37 & 5.41 & 1.00 & 2.00 \\
AuroraCap~\cite{auroracap}       & 31.57 & 31.56 & 27.88 & 5.00 & 21.02 & 26.01 & 5.00 & 10.07 \\
ShareGPT4Video~\cite{sharegpt4video}  & 30.19 & 32.06 & 11.93 & 2.53 & 7.64  & 9.18  & 2.48 & 5.26 \\
LLaVA-Video~\cite{llava_video}     & 79.34 & 81.08 & 27.88 & 6.00 & 22.34 & 27.56 & 6.00 & 10.46 \\
Vript~\cite{vript}           & 42.90 & 59.06 & 6.69  & 3.00 & 4.54  & 4.50  & 3.00 & 8.21 \\
GLaVE-Cap (Ours)       & 30.19 & 32.06 & 11.93 & 2.53 & 7.64  & 9.18  & 2.48 & 5.26 \\
\bottomrule
\end{tabular}
}
\vspace{1em}
\end{table}

\subsection{Evaluation Stability}
In \cref{sec:benchmark}, we introduce an automatic evaluation strategy for GLaVE-Bench, where GPT-4o serves as the judge model to select the best answer based on the captions under evaluation. However, since GPT-4o may exhibit instability in its judgment, this section focuses on analyzing the stability of the evaluation process. 
Specifically, we conduct each evaluation three times for every method and assess the consistency of the results. As shown in \cref{tab:consistency}, the calculated metrics exhibit minimal fluctuation. Furthermore, we categorize each question into three types based on the consistency of the results:
\begin{itemize}
    \item \textbf{Consistent}: All three evaluations yield the same result. This indicates a reliable and stable judgment by GPT-4o. We further divide this type into three sub-categories: ``correct'', ``wrong'', and ``not mention''.
    \item \textbf{Inconsistent due to `E'}: The three results are not identical, but the variation only occurs between the label ``E'' (i.e., Not Mentioned) and one specific non-``E'' option. This typically happens when the required information is only implicitly stated or must be inferred from the caption. The boundary between selecting ``E'' or a concrete answer is often ambiguous, leading to this kind of inconsistency. Despite this, such evaluations are still reliable to some extent, and has a reasonable influence to the metrics.
    \item \textbf{Fully Inconsistent}: The results differ across more than two non-``E'' options, reflecting a higher degree of instability in the evaluation process.
\end{itemize}

\begin{table*}[!t]
  \caption{Results on GLaVE-Bench across three runs.}
  \centering
  \resizebox{0.99\linewidth}{!}{
  \begin{tabular}{llccccccccccc}     
        \toprule
        \multicolumn{1}{l}{        \multirow{2}{*}{Method}} &
        \multicolumn{1}{l}{\multirow{2}{*}{V-L backbone}} &
        \multicolumn{3}{c}{Run 1} &
        \multicolumn{3}{c}{Run 2} &
        \multicolumn{3}{c}{Run 3}\\
        \cmidrule{3-11}
        \multicolumn{1}{l}{} & \multicolumn{1}{l}{} &
        Acc. & 
        Hall.$\downarrow$ &
        N.M.$\downarrow$ &
        Acc. & 
        Hall.$\downarrow$ &
        N.M.$\downarrow$ &
        Acc. & 
        Hall.$\downarrow$ &
        N.M.$\downarrow$\\
        \midrule
        \multicolumn{1}{l}{\multirow{2}{*}{LVD-2M~\cite{lvd2m}}} & Qwen2.5-VL & 41.73& 19.54 & 48.13 & 42.23 & 20.23 & 47.07 & 42.66 & 19.29 & 47.14  \\
        \multicolumn{1}{l}{} & GPT4o & 37.36 & 19.60 & 53.54 & 38.08 & 19.92 & 52.44 & 37.50 & 19.38 & 53.49  \\
        \midrule
        \multicolumn{1}{l}{\multirow{2}{*}{AuroraCap~\cite{auroracap}}} & Qwen2.5-VL &40.04 &  20.59& 49.58 &40.02 & 20.96 & 49.36 & 39.93 & 21.24 & 49.30 \\
        \multicolumn{1}{l}{} & GPT4o & 40.86 & 21.68 & 47.84 & 41.46 & 21.50 & 47.19 & 41.18 & 21.96 & 47.23   \\
        \midrule
        \multicolumn{1}{l}{\multirow{2}{*}{ShareGPT4Video~\cite{sharegpt4video}}} & Qwen2.5-VL & 47.93 & 19.74 & 40.29 & 47.90 & 20.06 & 40.09 & 48.08 & 19.08 & 40.58  \\
        \multicolumn{1}{l}{} & GPT4o & 45.39 & 18.23 & 44.49 & 45.49 & 17.88 & 44.60 & 46.03 & 17.44 & 44.25  \\
        \midrule
        \multicolumn{1}{l}{\multirow{2}{*}{LLaVA-Video~\cite{llava_video}}} &Qwen2.5-VL & 46.63 & 19.26 & 42.24 & 46.48 & 20.06 & 41.86 & 46.56 & 19.63 & 42.07  \\
        \multicolumn{1}{l}{} & GPT4o & 47.57 & 19.77 & 40.70 & 48.04 & 18.76 & 40.87 & 48.02 & 18.53 & 41.06  \\
        \midrule
        \multicolumn{1}{l}{\multirow{2}{*}{Vript~\cite{vript}}} & Qwen2.5-VL & 50.56 & 19.26 & 37.37 & 51.21 & 18.75 & 36.97 & 51.04 & 19.06 & 36.94   \\
        \multicolumn{1}{l}{} & GPT4o & 55.28 & 18.77 & 31.95 & 55.42 & 18.55 & 31.97 & 55.25 & 18.43 & 32.28  \\
        \midrule
        \midrule
        \multicolumn{1}{l}{\multirow{2}{*}{GLaVE-Cap (Ours)}} & Qwen2.5-VL&63.93 & 17.03 & 22.94 & 63.26 & 17.90 & 22.95 & 63.83 & 17.32 &22.80   \\ 
        \multicolumn{1}{l}{} & GPT4o& 67.68 & 16.00 & 19.43  &67.69  & 16.02 &19.40 &67.66 &16.09 &19.37   \\
        \bottomrule
  \end{tabular}
  }
  \label{tab:consistency}
\end{table*}

\begin{figure}[t]
  \centering
  \includegraphics[width=1.0\textwidth]{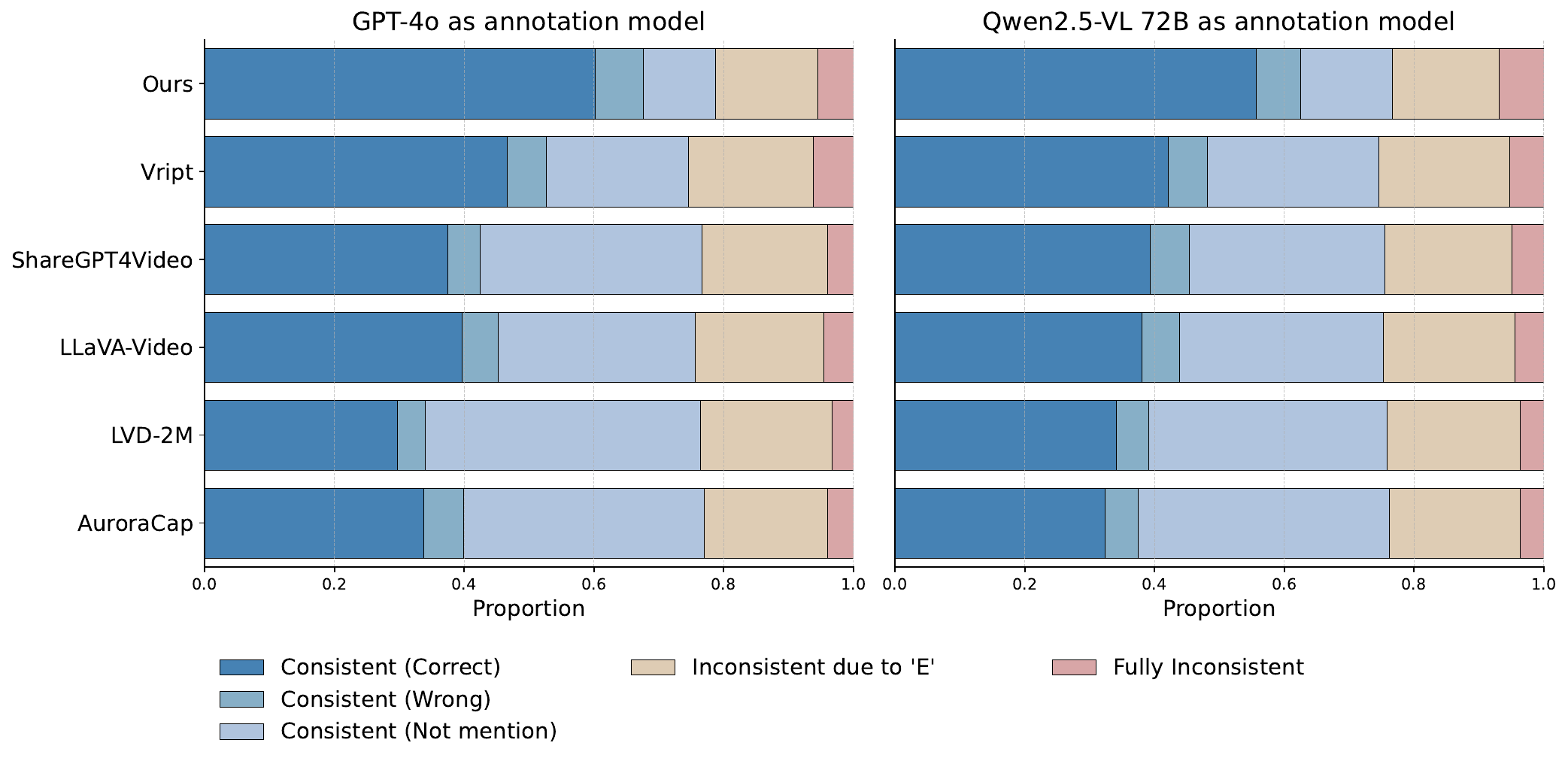}
  \caption{Question consistency type distribution across different methods and base models}
  \label{fig:consistency}
\end{figure}
We visualize the proportion of each category in \cref{fig:consistency}. It can be shown that most of the evaluations fall into the \textbf{Consistent} category, indicating that GPT-4o generally provides stable judgments. Notably, the accuracy improvement achieved by our method is primarily attributed to the ``Consistent'' category -- Our method yields confident and consistent ``correct'' judgments in 10\%-20\% questions that are answered `E' by other methods. This confirms that our performance gain is a reliable reflection of the caption quality.

\section{Training Details}\label{appendix:Training-details}
All models, including GLaVE-7B and those fine-tuned on other datasets, are trained based on the Qwen2.5-VL-7B model~\cite{bai2025qwen2}. The training data for each model consists of two parts: image data and video data.
The image data is identical across all models, comprising 405,752 general-category samples from LLaVA-One-Vision~\cite{li2024llava}. The video data, on the other hand, differs depending on the specific dataset used for fine-tuning. The detailed composition of the video data for each model is provided below:
\begin{itemize}
  \item \textbf{Vript}: A total of 408,816 video-caption pairs, including 8,776 samples from short videos and 400,040 from long videos.
  \item \textbf{ShareGPT4Video}: Contains 255K samples, including 240,000 open-ended QA items and 15,000 caption entries. The subset selection follows the same criteria as used in the LLaVA-Video model’s \texttt{llava-hound} training set.
  \item \textbf{LLaVA-Video-178K}: Composed of 5 sources, identical to the training set of the LLaVA-Video model:
  \begin{itemize}
    \item LLaVA-Video-178: 178,510 caption entries, including 960,792 open-ended QA items and 196,198 multiple-choice QA items 
    \item NeXT-QA: 17,090 open-ended QA items and 17,024 multiple-choice QA items
    \item ActivityNetQA: 23,530 open-ended QA items
    \item PerceptionTest: 1,803 open-ended QA items
    \item LLaVA-Hound: 240,000 open-ended QA items and 15,000 caption entries.
  \end{itemize}
  \item \textbf{GLaVE-1.2M}: Consists of 1,146,818 open-ended QA items. Caption data is excluded from training due to its small volume and excessive length, which may negatively impact learning.
\end{itemize}
Training is performed on 64 Ascend 910B2 devices distributed across 8 nodes (8 devices per node, each has 64G memory), using the MindSpeed-MM framework. We follow the official training configuration provided in the MindSpeed-MM repository\footnote{\url{https://gitee.com/ascend/MindSpeed-MM/blob/master/examples/qwen2.5vl/finetune_qwen2_5_vl_7b.sh}}. The only change is that we set GRAD\_ACC\_STEP to 16, resulting in a global batch size of 1,024. The training was conducted for a total of 2,000 iterations, taking approximately 60 hours to complete.

\section{Ablation Settings}\label{appendix:Ablation}
The detailed ablation settings are as follows:
\begin{itemize}
    \item \textbf{w/o visual prompt}: We replace the visual-masked inputs $\mathcal{M}_{i-1}$ and $\mathcal{M}_i$ in Eq.~\ref {DFC},~\ref{DTC} with the original keyframes $\mathcal{K}_{i-1}$ and $\mathcal{K}_i$ and remove textual supplementary information $S$ from the input. We also remove all phrases that are related to visual prompts and textual supplementary information in the corresponding prompts. The rest of the architecture remains unchanged.
    \item \textbf{w/o overview caption}: We remove the overview caption input from Eq.~\ref{DFC}-\ref{MDC},~\ref{SC}, along with all the overview-related phrases in the corresponding prompts. The rest of the architecture remains unchanged.
    \item \textbf{w/o dual-stream structure}: We directly use the differential caption $C_\texttt{diff}^i$ as the local caption $C_\texttt{local}^i$, without merging it with the detailed caption $C_\texttt{detail}^i$. The subsequent scene-level summarization strategy remains unchanged.
    \item \textbf{w/o adaptive scene-split}: We treat the entire video as a single scene, \ie, $len(\mathcal{SS}) = 1$ with $\mathcal{SS}_1.\texttt{ST} = 1$ and $\mathcal{SS}_1.\texttt{ED} = n$, where $n$ is the number of keyframes. Then, we perform the summary using the standard scene-level summarization approach. The local caption generation strategy remains unchanged. 
\end{itemize}

\section{Direct evaluation of module effects}\label{appendix:direct_ablation}
We conduct further analyses on the generated captions for all 55 videos of \textit{GLaVE-Bench}.

\subsection{TrackFusion addresses insufficient object attribute and relation description.}
To demonstrate this, we compare GLaVE-Cap with an ablated version that removes the dual-stream structure, by leveraging GPT-4o to count the number of direct visual description words (e.g., attributes and interactions of scenes, objects, and characters) in both local and video captions. The table below shows the average word counts, calculated across the 55 videos.

\begin{table}[h]
  \centering
  \caption{Average number of descriptive words in local and video captions.}
  \label{tab:trackfusion}
  \resizebox{0.99\linewidth}{!}{
  \begin{tabular}{lcccccccc}
    \toprule
    \multirow{2}{*}{Method} & \multicolumn{4}{c}{Local} & \multicolumn{4}{c}{Video} \\
    \cmidrule(lr){2-5} \cmidrule(lr){6-9}
     & \#Scene & \#Object & \#Person & Total & \#Scene & \#Object & \#Person & Total \\
    \midrule
    w/o Dual-stream & 4402 & 4924 & 4905 & 14229 & 41 & 38 & 33 & 113 \\
    GLaVE-Cap       & 6344 & 6720 & 5987 & 19046 & 48 & 45 & 38 & 133 \\
    \bottomrule
  \end{tabular}
  }
\end{table}

GLaVE-Cap consistently produces significantly richer visual content. At the local caption level, it generates 30\% more descriptive words compared to the ablated version. The advantage remains evident at the video-caption level (+18\%), although some detail is inevitably compressed during summarization. These results demonstrate that the dual-stream structure enhances the model’s ability to capture fine-grained visual information.

\subsection{Overview Caption addresses poor contextual consistency.}
We compare ShareGPT4Video, GLaVE-Cap, and its ablated version that removes the overview caption injection, by asking GPT-4o to count the number of contextual inconsistencies across local captions and in the video captions. Inconsistencies include: (1) object inconsistency: a human/object transforming into an entirely different one; (2) attribute inconsistency: the same human/object abruptly changing in color, shape, etc., without any clear cause. The table below shows the average metrics across all 55 videos.

\begin{table}[h]
  \centering
  \caption{Contextual inconsistency across local and video captions (per 1k words).}
  \label{tab:overview}
  \begin{tabular}{lcc}
    \toprule
    Method & Local $\downarrow$ & Video $\downarrow$ \\
    \midrule
    ShareGPT4Video      & 0.375 & 0.429 \\
    w/o Overview Caption& 0.145 & 0.086 \\
    GLaVE-Cap           & 0.102 & 0.046 \\
    \bottomrule
  \end{tabular}
\end{table}

Even the ablated version leads to significantly higher local-global consistency than ShareGPT4Video, demonstrating our TrackFusion design enables more detailed local captions that benefit the following summary stages in terms of ensuring contextual consistency. Incorporating the overview caption brings further improvement. We further manually inspect all video captions and 100 local captions. GPT-4o misidentified fewer than 5\% of inconsistencies, suggesting the reliability of the LLM-based analysis. Notably, in our caption, inconsistencies rarely occur in key elements such as people and objects, since they are typically covered by the overview caption.

\subsection{Adaptive Scene-split addresses information loss during summarization.}
We compare GLaVE-Cap and its ablated version that removes the adaptive scene-split module. Specifically, we ask GPT-4o to check whether each visual description in the local captions is preserved in the final video-level caption. We report the average of this proportion, calculated across all local captions from the 55 videos, as shown below.

\begin{table}[h]
  \centering
  \caption{Proportion of preserved visual descriptions in final video captions.}
  \label{tab:scene-split}
  \begin{tabular}{lcccc}
    \toprule
    Method & Scene & Object & Person & Average \\
    \midrule
    w/o Adaptive Scene-split & 72.09 & 57.33 & 58.61 & 62.89 \\
    GLaVE-Cap                & 90.80 & 81.80 & 77.07 & 83.77 \\
    \bottomrule
  \end{tabular}
\end{table}

Incorporating adaptive scene-split leads to a clear improvement in retaining detailed information (+20\%). These results confirm that the adaptive scene-split effectively reduces information loss during summarization, thereby producing more comprehensive and faithful video captions.

\section{Impact of Prompt Design and Base Model Capability}\label{appendix:prompt_model}
\subsection{Analysis of how different types of prompts affect model performance}
The prompt design plays a vital role in the performance of GLaVE-Cap. Our original prompts are based on a hierarchical structured design, inspired by prior works such as ShareGPT4Video and LLaVA-Video. We further investigate how different prompt templates affect the performance of the annotation model.

We focus on two subsets of the prompts used in GLaVE-Cap: those crucial for TrackFusion to extract visual details for local captions, and those crucial for CaptionBridge to summarize local captions into global captions. We choose two alternative prompt styles for evaluation:

\begin{itemize}
  \item \textbf{Simplified Version:} Prompts retain only necessary parts to describe the task, input, and output, while removing specific guidance (e.g., constraints, skills, enumerated visual aspects).
  \item \textbf{Unstructured Version:} Prompts are rewritten by GPT-4o into a natural language paragraph without subtitles or lists, while preserving semantic content.
\end{itemize}

To assess the impact of different prompt types, we modify one prompt subset at a time and keep the remaining prompts unchanged. All experiments are conducted using GPT-4o as the annotation model. Due to API cost, we ran the experiment on a randomly sampled subset of \textit{GLaVE-Bench}, which includes 10 videos and 1,090 questions. Accuracy results are shown below:

\begin{table}[h]
  \centering
  \caption{Effect of different prompt styles on GLaVE-Cap performance.}
  \label{tab:prompt}
  \begin{tabular}{lcc}
    \toprule
    Version & Local Caption & Global Caption \\
    \midrule
    Original Version    & 69.85 & 69.85 \\
    Simplified Version  & 68.01 & 63.39 \\
    Unstructured Version& 69.57 & 72.35 \\
    \bottomrule
  \end{tabular}
\end{table}

The results show that simplified prompts significantly degrade caption quality, suggesting that task-specific guidance is crucial for eliciting high-quality responses. Interestingly, unstructured prompts lead to similar or even better performance (+2.5 for global captions). This reveals that GLaVE-Cap may further benefit from prompt optimization strategies. The results also suggest that global caption generation is more sensitive to prompt design.

\subsection{Analysis of how the capabilities of (M)LLMs affect video captioning}
To better understand how the capabilities of (M)LLMs affect video captioning, we conducted additional experiments. Beyond the results already reported for GPT-4o and Qwen2.5-VL-72B, we further evaluated two smaller models, Qwen2.5-VL-32B and Qwen2.5-VL-7B, on \textit{GLaVE-Bench}. Notably, Qwen2.5-VL-7B struggles to understand the prompt and fails to complete the task, often producing repetitive patterns and unreliable structured output. Qwen2.5-VL-32B performs relatively better but still shows limitations, such as failing to provide valid scene segmentation or generating captions in structured formats despite explicit instructions. Quantitative results are shown below:

\begin{table}[h]
  \centering
  \caption{Performance of different (M)LLMs on \textit{GLaVE-Bench}.}
  \label{tab:mlmms}
  \begin{tabular}{lccc}
    \toprule
    Model & Acc $\uparrow$ & Hall $\downarrow$ & N.M. $\downarrow$ \\
    \midrule
    GPT-4o          & 67.68 & 16.04 & 19.40 \\
    Qwen2.5-VL-72B  & 63.67 & 17.42 & 22.90 \\
    Qwen2.5-VL-32B  & 61.05 & 18.74 & 24.87 \\
    Qwen2.5-VL-7B   & N/A   &   N/A  &   N/A \\
    \bottomrule
  \end{tabular}
\end{table}

We further explore the influence of the annotation model in individual modules of GLaVE-Cap, including TrackFusion and the summary step of CaptionBridge. We use GPT-4o as the main annotation model, but replace those modules with Qwen2.5-VL-32B. Results are shown below:

\begin{table}[h]
  \centering
  \caption{Module-wise replacement analysis on \textit{GLaVE-Bench}.}
  \label{tab:module}
  \begin{tabular}{lccc}
    \toprule
    Setting & Acc $\uparrow$ & Hall $\downarrow$ & N.M. $\downarrow$ \\
    \midrule
    All GPT-4o               & 67.68 & 16.04 & 19.40 \\
    Qwen replace TrackFusion  & 58.34 & 18.42 & 28.49 \\
    Qwen replace Summary Step & 65.08 & 16.66 & 21.91 \\
    \bottomrule
  \end{tabular}
\end{table}

These results indicate that local caption generation (TrackFusion) relies more heavily on powerful annotation models, while the summarization step shows relatively weaker dependence.

\section{More ablation studies}\label{appendix:more_ablation}
We have conducted ablation studies on both \textit{VidCapBench}~\cite{chen2025vidcapbench} and the short subset of \textit{Video-MME}~\cite{videomme} (Video-MME-S). The settings are identical to Table~\ref{tab-ablation}, but use Qwen2.5-VL-72B as the annotation model due to the prohibitive cost of GPT-4o.

\begin{table}[h]
  \centering
  \caption{Ablation study on \textit{VidCapBench} and \textit{Video-MME-S}.}
  \label{tab:ablation_other_benchmarks}
  \resizebox{0.99\linewidth}{!}{
  \begin{tabular}{lcccc}
    \toprule
    Method & VidCapBench-1 & VidCapBench-2 & VidCapBench-3 & Video-MME-S \\
    \midrule
    w/o Adaptive Scene-split   & 16.52 & 56.50 & 88.69 & 65.11 \\
    w/o Dual-stream Structure  & 17.13 & 57.59 & 89.05 & 70.78 \\
    w/o Overview Caption       & 18.58 & 58.39 & 92.41 & 72.67 \\
    w/o Visual Prompt          & 18.96 & 58.57 & 92.49 & 73.00 \\
    \textbf{Ours}              & 18.78 & 60.98 & 92.34 & 73.22 \\
    \bottomrule
  \end{tabular}
  }
\end{table}

The observed trends are generally consistent with those on GLaVE-Bench. However, the improvements are less pronounced, largely because \textit{VidCapBench} consists mainly of short ($\approx$10s) clips, and \textit{Video-MME} focuses on video reasoning rather than comprehensive and fine-grained video understanding.

Additionally, in \cref{appendix:direct_ablation}, we provide a GPT-4o-based quantitative analysis on object attribute and relation description, contextual consistency, and summarization quality, demonstrating the effectiveness of our proposed modules.

These results suggest that the modest gains on existing benchmarks stem from their limited capacity to capture the types of fine-grained improvements our method offers, highlighting the motivation and necessity for GLaVE-Bench.

\section{Hallucination Accumulation Analysis}\label{appendix:Hallucination}
Video detailed captioning typically involves multiple steps~\cite{llava_video,sharegpt4video}, and our framework follows a similar multi-stage process. However, most steps in our framework are pure-text summarization processes, which significantly limit the propagation of hallucinations. Notably, GPT-4o and Qwen2.5-72B, the backbone we adopted, achieve only 1.5\% and 4.3\% hallucination rate on the Hughes Hallucination Evaluation Model (HHEM) leaderboard\footnote{\url{https://huggingface.co/spaces/vectara/leaderboard}}, indicating high reliability for text-only tasks. The main source of potential hallucination arises from extracting dynamic changes and static details from keyframes, which inherently involves vision-language interaction and is difficult to avoid in the video detailed captioning task. To reduce hallucination in this stage, we have adopted visual prompts provided by expert models. This strategy has been shown to be effective in minimizing hallucinations in prior works such as Set-of-Mark~\cite{som}.

To further evaluate potential hallucination accumulation, we conduct a quantitative evaluation on the semantic detail subset of \textit{VideoHallucer}\cite{VideoHallucer}. We follow the caption evaluation protocol introduced in the main paper (same as Table~\ref{tab-main-caption}). Our method achieves a higher positive accuracy while maintaining hallucination accuracy comparable to baseline methods (see Table~\ref{tab:hallucination}). It demonstrates that our pipeline does not suffer from noticeable hallucination accumulation.

\begin{table}[h]
  \centering
  \caption{Comparison on the semantic detail subset of \textit{VideoHallucer}.}
  \label{tab:hallucination}
  \begin{tabular}{lccc}
    \toprule
    Method & Basic $\uparrow$ & Hallucinated $\downarrow$ & Overall $\uparrow$ \\
    \midrule
    LVD-2M         & 52.0 & 96.0 & 49.0 \\
    AuroraCap      & 41.0 & 93.0 & 35.5 \\
    ShareGPT4Video & 57.0 & 93.0 & 52.0 \\
    LLaVA-Video    & 59.5 & 94.5 & 55.5 \\
    Vript          & 60.5 & 95.0 & 58.0 \\
    \textbf{Ours}  & 62.0 & 93.5 & 56.5 \\
    \bottomrule
  \end{tabular}
\end{table}

\section{Showcase Caption}\label{appendix:caption_show}
\begin{figure}[htbp]
  \centering
  \includegraphics[width=1.0\textwidth]{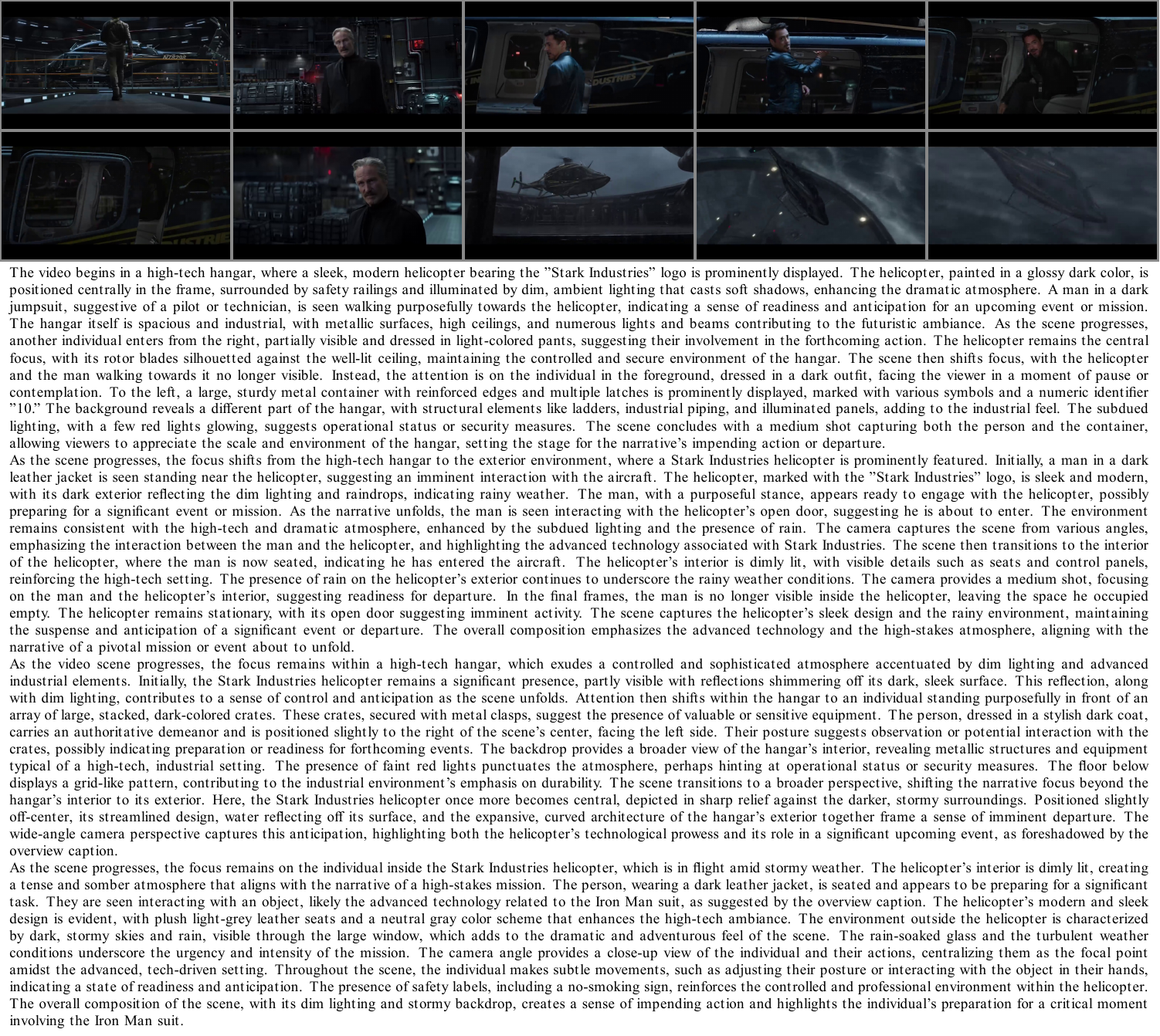}
  \caption{The first part of the showcase caption}
  \label{fig:caption_show1}
\end{figure}

\begin{figure}[htbp]
  \centering
  \includegraphics[width=1.0\textwidth]{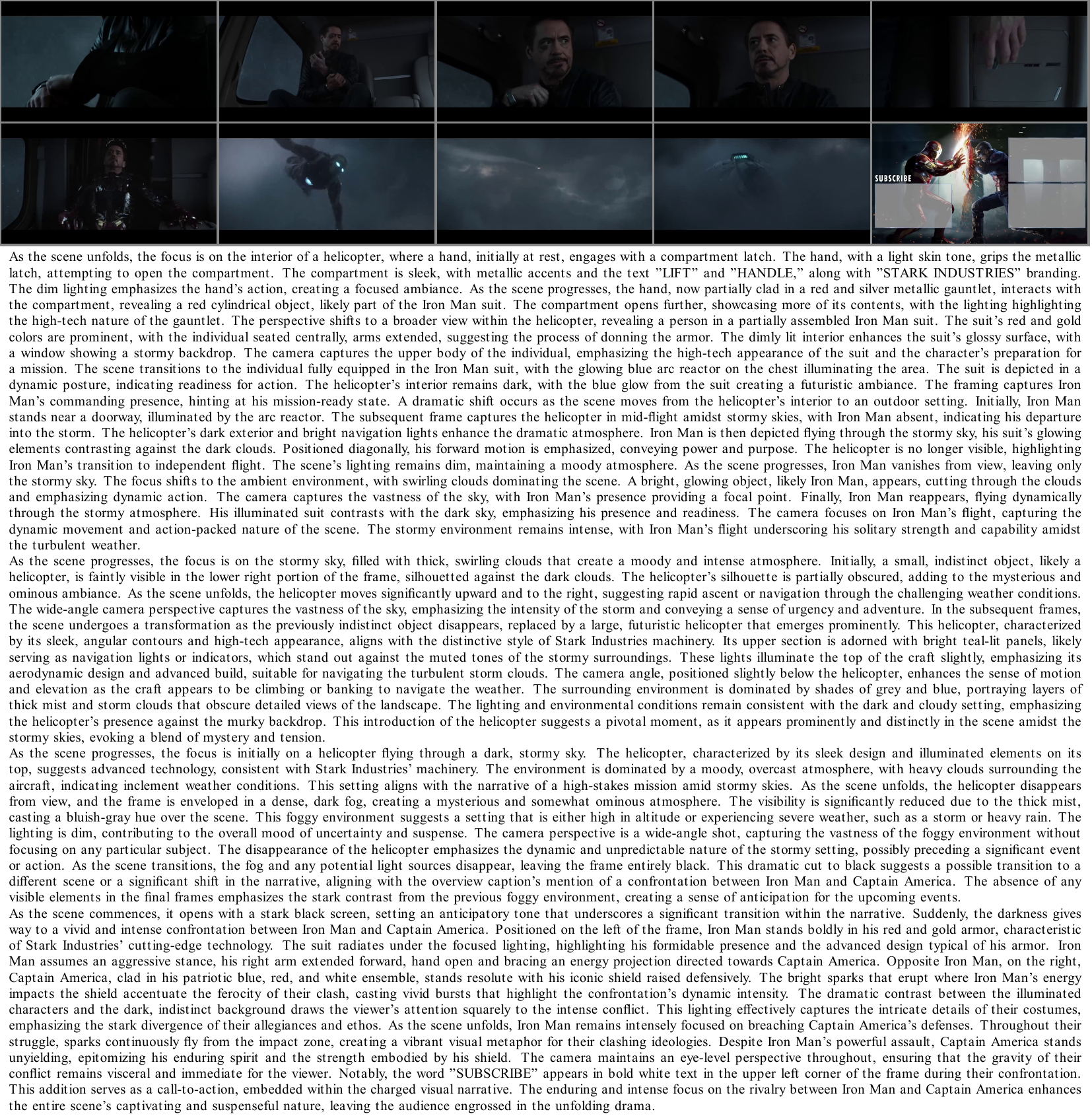}
  \caption{The second part of the showcase caption}
  \label{fig:caption_show2}
\end{figure}

In this section, we showcase a caption generated in GLaVE-Cap in \cref{fig:caption_show1,fig:caption_show2}.

\section{GLaVE-Cap Framework Prompt Template}\label{appendix:frame_prompt}
\begin{figure}[htbp]
  \centering
  \includegraphics[width=1.0\textwidth]{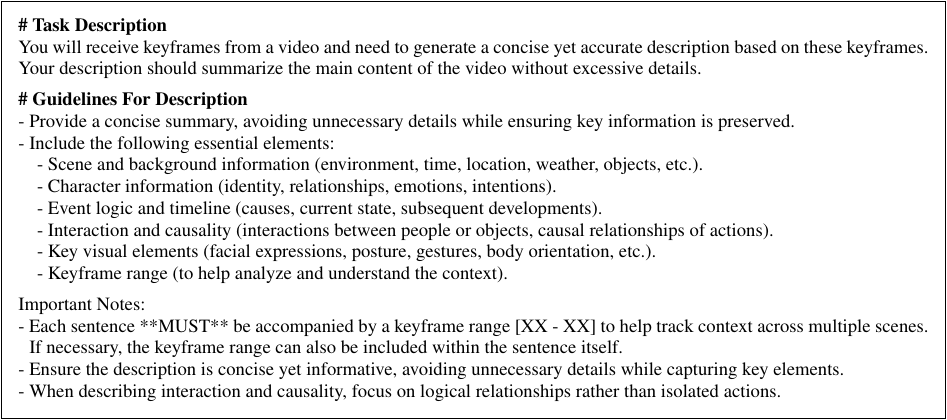}
  \caption{$\texttt{prompt}_\texttt{overview}$}
  \label{fig:overview}
\end{figure}

\begin{figure}[htbp]
  \begin{center}
  \includegraphics[width=1.0\textwidth]{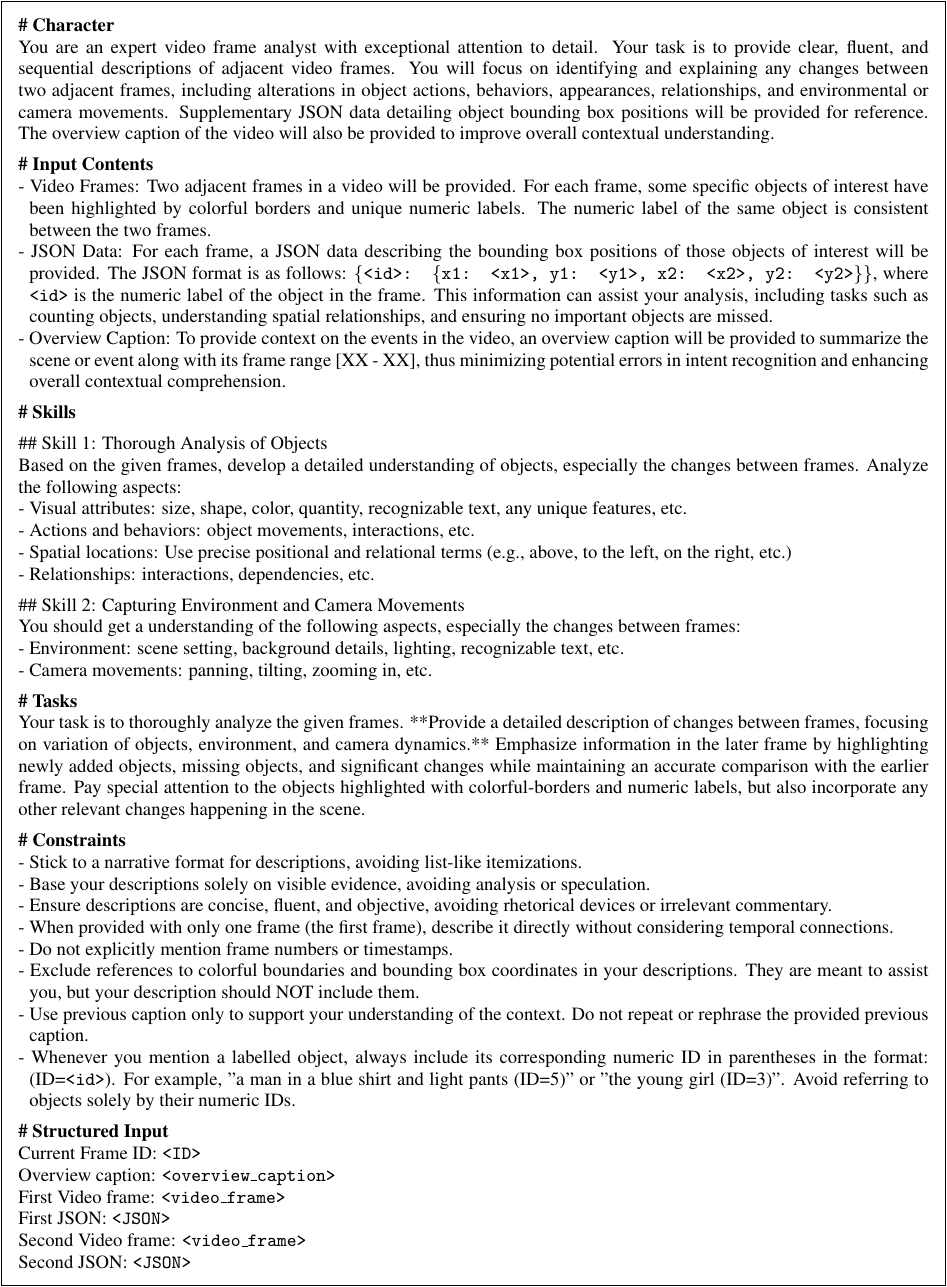}
  \end{center}
  \caption{$\texttt{prompt}_\texttt{diff}$}
  \label{fig:different}
\end{figure}

\begin{figure}[htbp]
  \centering
  \begin{center}
  \includegraphics[width=1.0\textwidth]{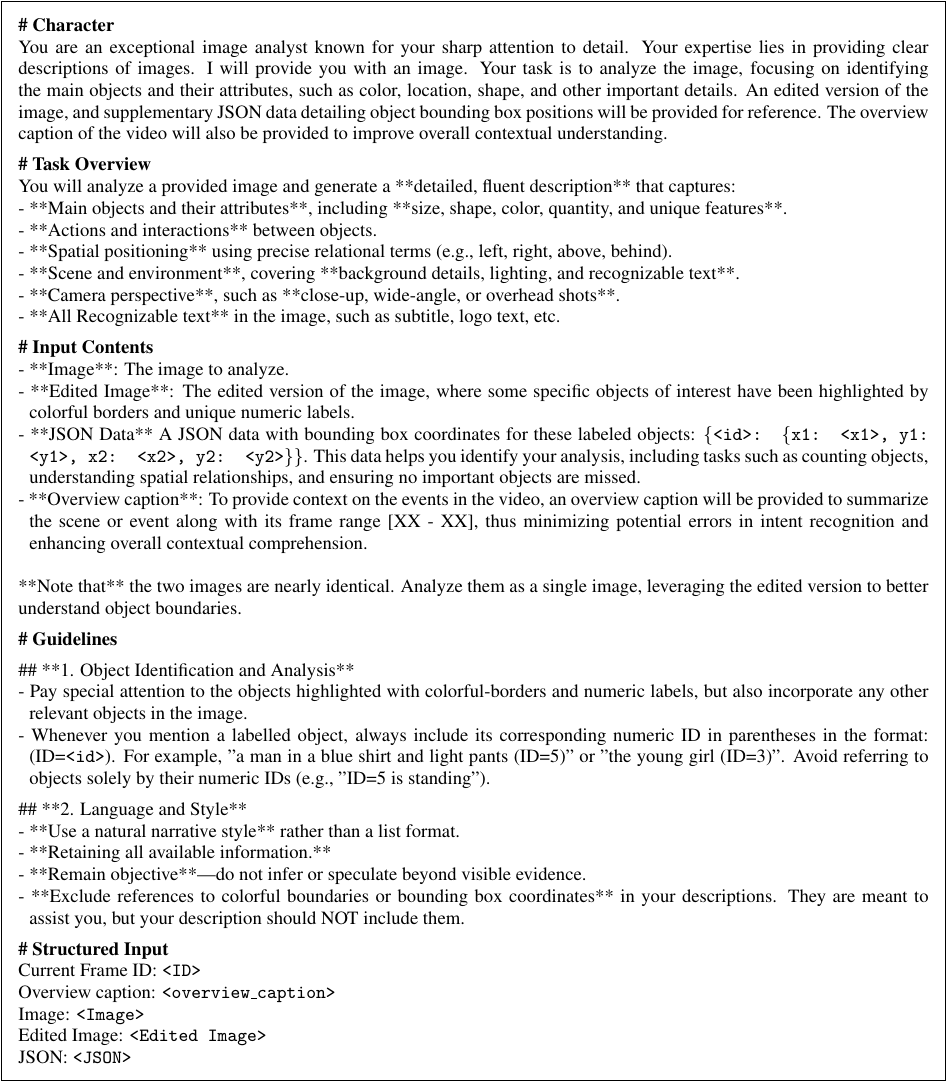}
  \end{center}
  \caption{$\texttt{prompt}_\texttt{detail}$}
  \label{fig:detail}
\end{figure}

\begin{figure}[htbp]
  \centering
  \includegraphics[width=1.0\textwidth]{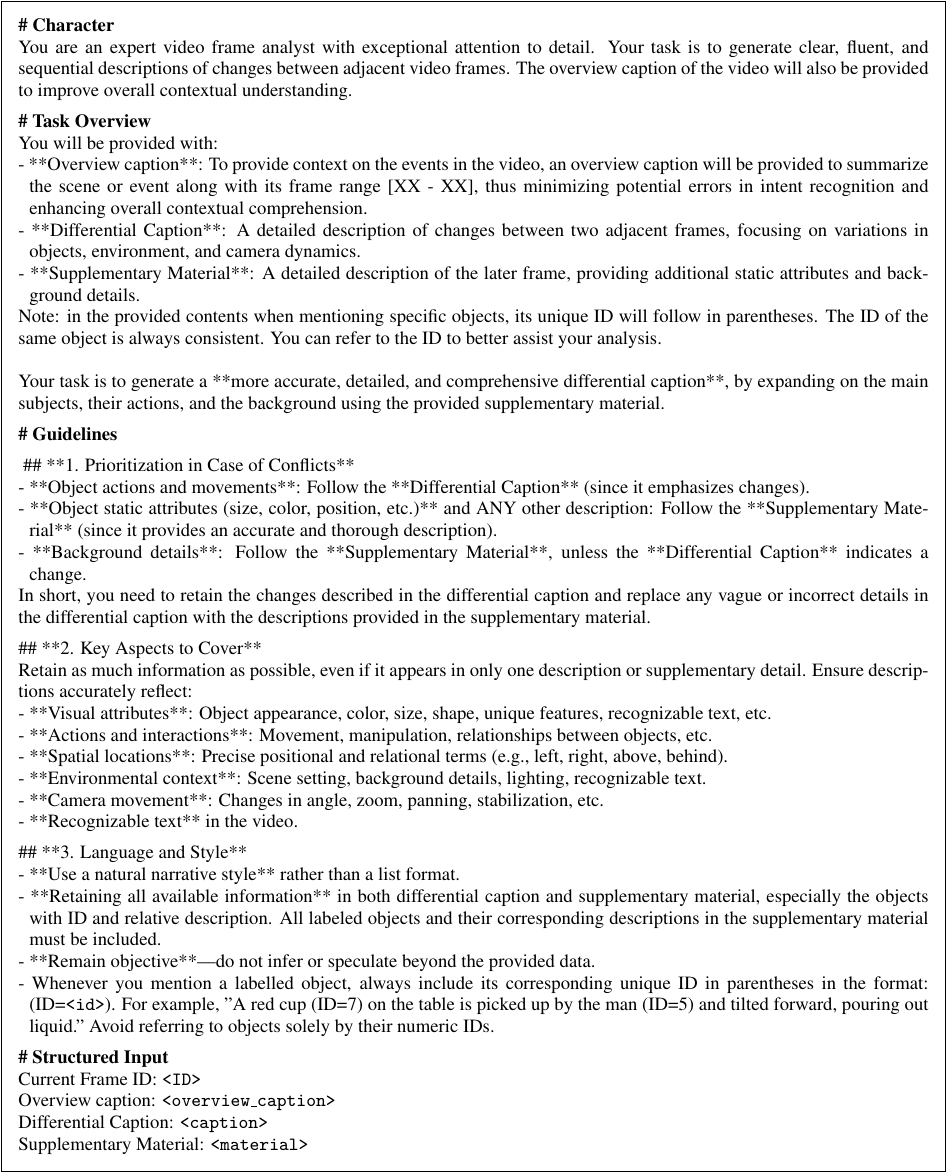}
  \caption{$\texttt{prompt}_\texttt{merge}$}
  \label{fig:merge}
\end{figure}

\begin{figure}[htbp]
  \centering
  \includegraphics[width=1.0\textwidth]{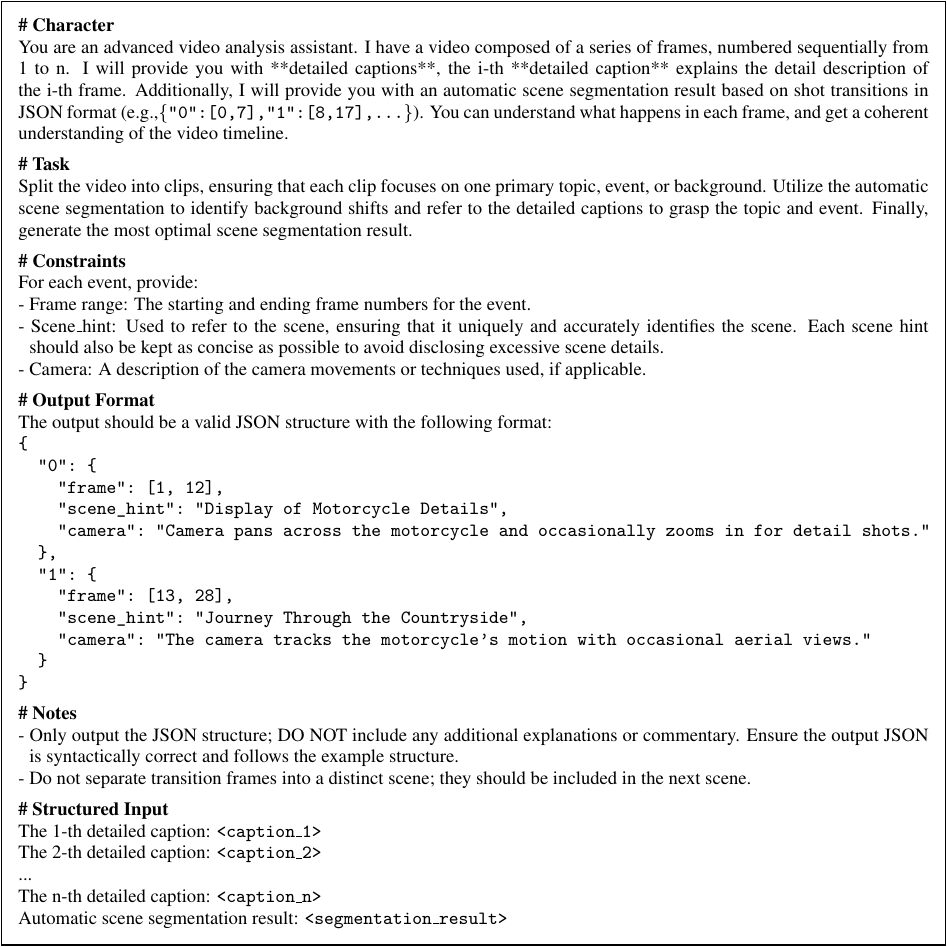}
  \caption{$\texttt{prompt}_\texttt{SS}$}
  \label{fig:split}
\end{figure}

\begin{figure}[htbp]
  \centering
  \includegraphics[width=1.0\textwidth]{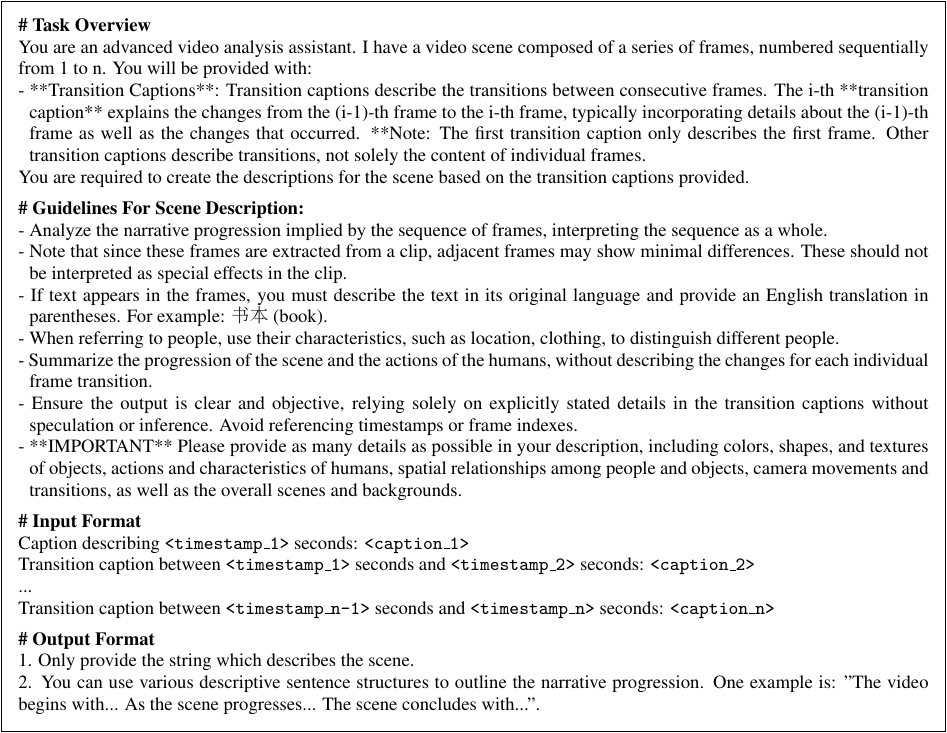}
  \caption{$\texttt{prompt}_\texttt{SC}$ for the first scene.}
  \label{fig:summary_first}
\end{figure}

\begin{figure}[htbp]
  \centering
  \includegraphics[width=1.0\textwidth]{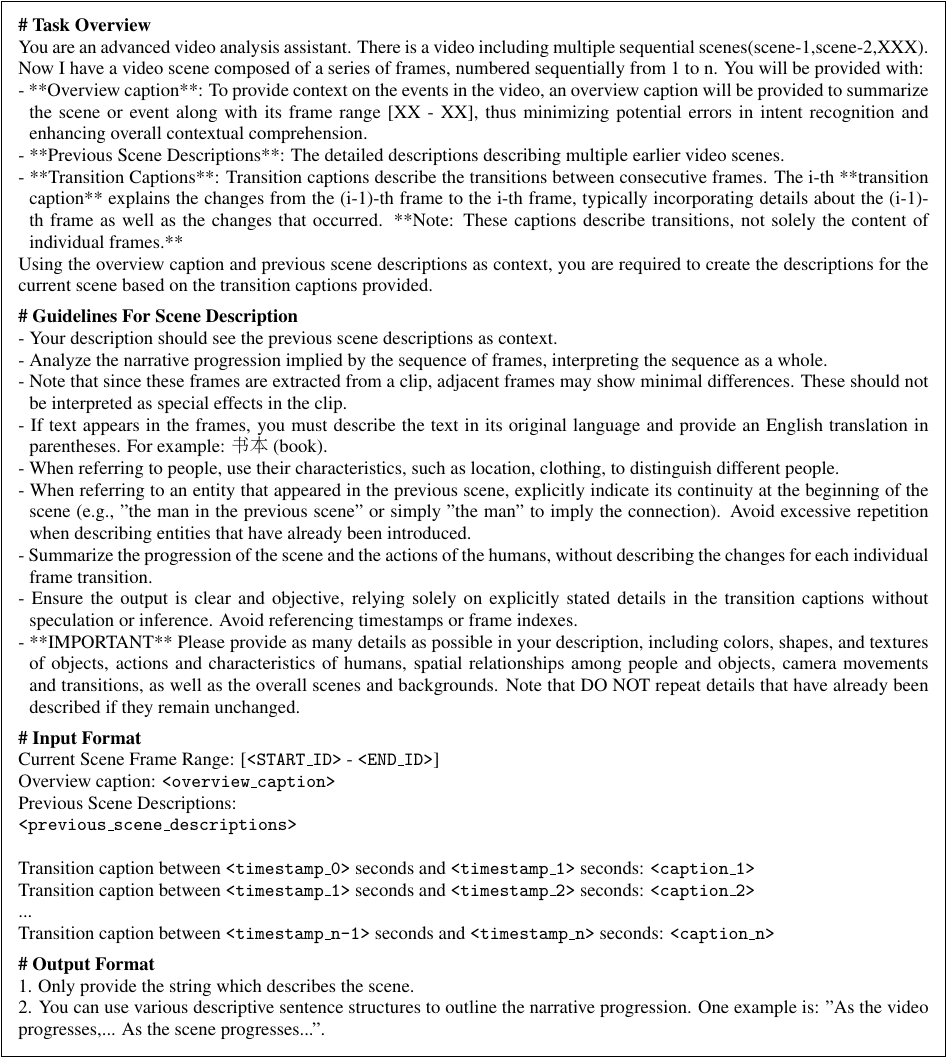}
  \caption{$\texttt{prompt}_\texttt{SC}$ for other scenes.}
  \label{fig:summary_second}
\end{figure}

In this section, we showcase the prompts used in GLaVE-Cap as \cref{fig:overview,fig:different,fig:detail,fig:merge,fig:split,fig:summary_first,fig:summary_second}.

\section{Benchmark Evaluation Prompt Template}\label{appendix:eval_prompt}
\begin{figure}[htbp]
  \centering
  \includegraphics[width=1.0\textwidth]{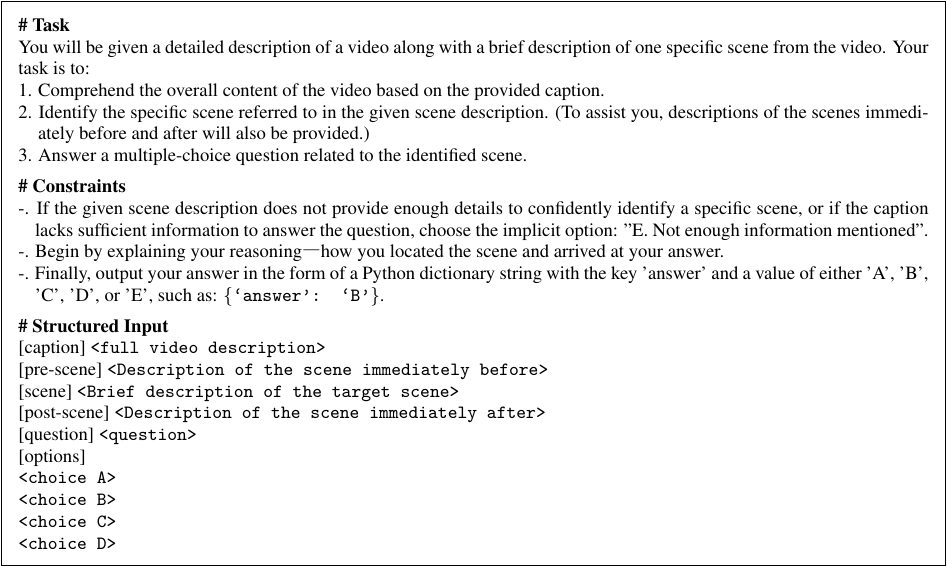}
  \caption{Prompt for GLaVE-Bench evaluation}
  \label{fig:eval_prompt}
\end{figure}

We have introduced an automatic evaluation strategy for GLaVE-Bench in \cref{sec:benchmark}. The prompt template used is illustrated in \cref{fig:eval_prompt}.

\section{Limitation and Societal Impacts}\label{appendix:limitation}
\noindent\textbf{Limitation.}
Although our current framework leverages vision experts to assist the VLM in generating local captions, leading to clear improvements over using the VLM alone, it remains constrained by the VLM’s capacity to process complex visual inputs within a single inference. In scenes containing numerous objects, the framework may still fail to capture important visual elements due to this bottleneck. To address this limitation, we plan to shift from keyframe-based to object-centric local captioning, enabling accurate object identification and comprehensive scene description through object tracking and multi-round interactions between vision experts and the VLM.

\noindent\textbf{Societal Impacts.}
1) Our dataset and benchmark annotations will be released publicly, contributing to the research community by facilitating progress in fine-grained video understanding and enabling more comprehensive and accurate evaluation of model capabilities.
2) Although the video data we use in GLaVE-Bench and GLaVE-1.2M is sourced from publicly available datasets, we cannot guarantee the complete absence of personally identifiable information, such as human faces. Users are therefore responsible for ensuring compliance with the licensing and ethical standards of the original video sources when using any part of the visual content.
3) Similar to other VLMs, our GLaVE-7B may be vulnerable to adversarial prompting that results in harmful or inappropriate outputs. This highlights the ongoing need for advances in AI safety to ensure responsible and secure use of such models.

\section{License}\label{appendix:license}
We will introduce licenses of our assets and the sources, usage, and licenses of existing assets used in this work:

\begin{itemize}
    \item \textbf{Code: Apache 2.0}
    \begin{itemize}
        \item \textbf{PySceneDetect} v0.6.6: Used for scene detection. URL: \url{https://github.com/Breakthrough/PySceneDetect}. Licensed under the \textbf{BSD-3-Clause License}.
        \item \textbf{Grounded-SAM2}: Used for object detection and segmentation. URL: \url{https://github.com/IDEA-Research/Grounded-Segment-Anything}. Licensed under the \textbf{Apache License 2.0}.
    \end{itemize}

    \item \textbf{Test Benchmark: Video-MME License (Academic use only; no commercial use; no redistribution)}
    \begin{itemize}
        \item We use the video from \textbf{Video-MME}~\cite{videomme}. URL: \url{https://github.com/MME-Benchmarks/Video-MME}. The dataset follows the Video-MME license (Video-MME is only used for academic research. Commercial use in any form is prohibited. The copyright of all videos belongs to the video owners.).
    \end{itemize}

    \item \textbf{Training dataset: Apache 2.0}
    \begin{itemize}
    \item We use the video from \textbf{LLaVA-Video-Interleaved-178K}~\cite{llava_video}. URL: \url{https://huggingface.co/datasets/lmms-lab/LLaVA-Video-178K}. The dataset follows the \textbf{Apache License 2.0} license. We respect the license terms and provide attribution accordingly.
    \end{itemize}

    \item \textbf{Pretrained Model: Apache 2.0}
    \begin{itemize}
        \item We fine-tuned our model based on \textbf{Qwen2.5-VL-7B}~\cite{bai2025qwen2}, developed by Alibaba. URL: \url{https://github.com/QwenLM/Qwen2.5-VL}. The model is licensed under the \textbf{Apache License 2.0}. We comply with its terms for research use.
    \end{itemize}
\end{itemize}

\begin{figure}[htbp]
  \centering
  \includegraphics[width=1.0\textwidth]{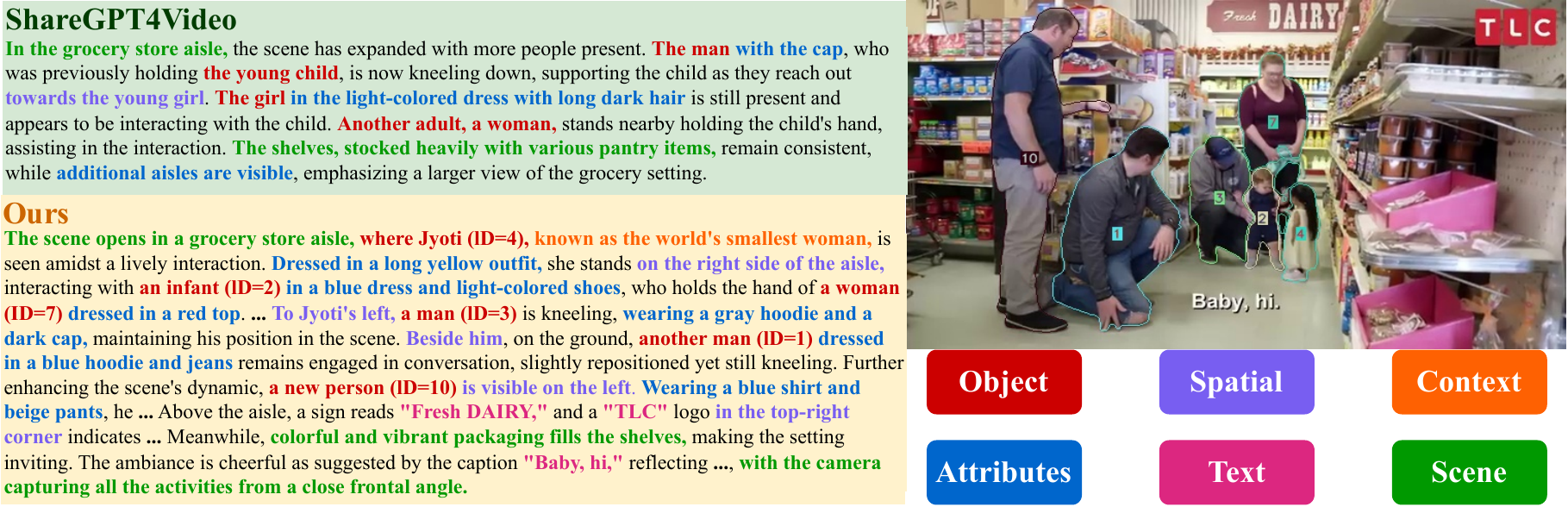}
  \caption{The full caption in \cref{fig:teaser}. We omit from our captions the parts not mentioned in \cref{fig:teaser}, primarily those describing actions, reasoning, and overall stylistic elements.}
  \label{fig:full_caption}
\end{figure}

\clearpage
\section*{NeurIPS Paper Checklist}

\begin{enumerate}

\item {\bf Claims}
    \item[] Question: Do the main claims made in the abstract and introduction accurately reflect the paper's contributions and scope?
    \item[] Answer: \answerYes{} 
    \item[] Justification: The abstract and introduction clearly state the main claims of the paper, including the proposal of a novel video-captioning framework, the construction of the evaluation benchmark and training dataset, and the experimental validation demonstrating the effectiveness of the approach. These claims accurately reflect the contributions and scope of the work and are supported by the results presented. Assumptions and limitations are acknowledged to provide appropriate context for the findings.

\item {\bf Limitations}
    \item[] Question: Does the paper discuss the limitations of the work performed by the authors?
    \item[] Answer: \answerYes{} 
    \item[] Justification: The paper discuss the limitations of the work in ~\cref{appendix:limitation}.

\item {\bf Theory Assumptions and Proofs}
    \item[] Question: For each theoretical result, does the paper provide the full set of assumptions and a complete (and correct) proof?
    \item[] Answer: \answerNA{} 
    \item[] Justification: The paper does not include theoretical results, such as theorems or formal proofs, as it focuses on methodological contributions and empirical validation.

    \item {\bf Experimental Result Reproducibility}
    \item[] Question: Does the paper fully disclose all the information needed to reproduce the main experimental results of the paper to the extent that it affects the main claims and/or conclusions of the paper (regardless of whether the code and data are provided or not)?
    \item[] Answer: \answerYes{} 
    \item[] Justification: The paper provides all necessary details to reproduce the main experimental results, including video-captioning method (\cref{sec:method}, \cref{sec:reproduction-detail}), prompts (\cref{appendix:frame_prompt}, \cref{appendix:eval_prompt}), evaluation protocol (\cref{sec:eval}), training setting (\cref{appendix:Training-details}), and ablation details (\cref{appendix:Ablation}).

\item {\bf Open access to data and code}
    \item[] Question: Does the paper provide open access to the data and code, with sufficient instructions to faithfully reproduce the main experimental results, as described in supplemental material?
    \item[] Answer: \answerYes{} 
    \item[] Justification: We will provide relevant code and data in the supplementary materials, including implementation of GLaVE-Cap, reproduced baseline methods, and GLaVE-Bench dataset as well as evaluation toolkit.

\item {\bf Experimental Setting/Details}
    \item[] Question: Does the paper specify all the training and test details (e.g., data splits, hyperparameters, how they were chosen, type of optimizer, etc.) necessary to understand the results?
    \item[] Answer: \answerYes{} 
    \item[] Justification: We provide relative information in \cref{sec:reproduction-detail,appendix:Training-details}.

\item {\bf Experiment Statistical Significance}
    \item[] Question: Does the paper report error bars suitably and correctly defined or other appropriate information about the statistical significance of the experiments?
    \item[] Answer: \answerNo{}{} 
    \item[] Justification: Due to the high computational cost, we did not conduct sufficient independent runs  to report error bars for evaluating and captioning, due to the high cost. Despite this, we report results over three independent runs and the analysis of experimental stability in \cref{appendix:Analysis}, thereby addressing the variability and robustness of the results.

\item {\bf Experiments Compute Resources}
    \item[] Question: For each experiment, does the paper provide sufficient information on the computer resources (type of compute workers, memory, time of execution) needed to reproduce the experiments?
    \item[] Answer: \answerYes{} 
    \item[] Justification: We report experiments compute resources in \cref{appendix:Training-details}.
    
\item {\bf Code Of Ethics}
    \item[] Question: Does the research conducted in the paper conform, in every respect, with the NeurIPS Code of Ethics \url{https://neurips.cc/public/EthicsGuidelines}?
    \item[] Answer: \answerYes{} 
    \item[] Justification: We carefully read the ethics review and comply with them.

\item {\bf Broader Impacts}
    \item[] Question: Does the paper discuss both potential positive societal impacts and negative societal impacts of the work performed?
    \item[] Answer: \answerYes{} 
    \item[] Justification: We report social impacts in \cref{appendix:limitation}.
    
\item {\bf Safeguards}
    \item[] Question: Does the paper describe safeguards that have been put in place for responsible release of data or models that have a high risk for misuse (e.g., pretrained language models, image generators, or scraped datasets)?
    \item[] Answer: \answerYes{} 
    \item[] Justification: The paper involves the release of models and data that could potentially be misused if applied irresponsibly. To prevent misuse, we provide a usage guideline that explicitly prohibits harmful or unethical use cases, such as generating misleading content or surveillance applications.

\item {\bf Licenses for existing assets}
    \item[] Question: Are the creators or original owners of assets (e.g., code, data, models), used in the paper, properly credited and are the license and terms of use explicitly mentioned and properly respected?
    \item[] Answer: \answerYes{} 
    \item[] Justification: We provide the license in \cref{appendix:license}.

\item {\bf New Assets}
    \item[] Question: Are new assets introduced in the paper well documented and is the documentation provided alongside the assets?
    \item[] Answer: \answerYes{} 
    \item[] Justification: The paper introduces new assets, including a dataset and pretrained models. We provide comprehensive documentation, including information about the data source, licensing, and the usage guidelines.

\item {\bf Crowdsourcing and Research with Human Subjects}
    \item[] Question: For crowdsourcing experiments and research with human subjects, does the paper include the full text of instructions given to participants and screenshots, if applicable, as well as details about compensation (if any)? 
    \item[] Answer: \answerYes{} 
    \item[] Justification: The benchmark was annotated by three volunteer annotators, with a total of approximately 230 hours spent on the annotation process. We will provide an annotation guideline in the supplementary material. 

\item {\bf Institutional Review Board (IRB) Approvals or Equivalent for Research with Human Subjects}
    \item[] Question: Does the paper describe potential risks incurred by study participants, whether such risks were disclosed to the subjects, and whether Institutional Review Board (IRB) approvals (or an equivalent approval/review based on the requirements of your country or institution) were obtained?
    \item[] Answer: \answerNo{} 
    \item[] Justification: Our study involved a small number of volunteer annotators who contributed to the benchmark annotation. The annotation task posed no foreseeable risks to participants, and all volunteers were informed about the nature and purpose of their contributions. Additionally, according to the regulations of our country, such annotation activities do not require IRB or equivalent approval.
    
\item {\bf Declaration of LLM usage}
    \item[] Question: Does the paper describe the usage of LLMs if it is an important, original, or non-standard component of the core methods in this research? Note that if the LLM is used only for writing, editing, or formatting purposes and does not impact the core methodology, scientific rigorousness, or originality of the research, declaration is not required.
    \item[] Answer: \answerYes{} 
    \item[] Justification:  We incorporated an LLM-based module to assist in generating synthetic training data for video QA. This component plays a non-trivial role in our method design and thus requires declaration.
\end{enumerate}
\end{document}